%% file: acl_latex.tex
\pdfoutput=1

\documentclass[11pt]{article}

\usepackage{latex/acl}

\usepackage{times}
\usepackage{latexsym}

\usepackage[T1]{fontenc}

\usepackage[utf8]{inputenc}

\usepackage{microtype}
\usepackage{dsfont}

\usepackage{inconsolata}

\input{commands.tex}
\usepackage{mathtools}

\title{Label-Efficient Model Selection for Text Generation}

\author{Shir Ashury-Tahan\thanks{\ \ These authors contributed equally to this work.}$^{\spadesuit}$$^{\heartsuit}$, Ariel Gera\footnotemark[1]$^{\spadesuit}$, Benjamin Sznajder$^{\spadesuit}$, \\
\textbf{Leshem Choshen$^{\spadesuit\diamondsuit}$, Liat Ein-Dor$^{\spadesuit}$ and Eyal Shnarch$^{\spadesuit}$}\\
$^{\spadesuit}$IBM Research, $^{\heartsuit}$Bar-Ilan University, $^{\diamondsuit}$MIT \\}

\begin{document}

\maketitle

\input{main}

\bibliography{references}

\appendix

\input{appendix}

\end{document}

%% file: commands.tex
\usepackage{graphicx} 
\usepackage{xcolor}
\usepackage{caption}
\usepackage[export]{adjustbox}
\usepackage{booktabs}
\usepackage[ruled,vlined]{algorithm2e}
\usepackage{hyperref}
\usepackage{amsmath}
\usepackage{enumitem}
\usepackage{multirow}
\usepackage{subfig}
\usepackage{fdsymbol}

\newcommand\ag[1]{\textcolor{teal}{[AG: #1]}}

\newcommand\method[0]{\textit{DiffUse}}

\newcommand{\cready}[1]{{}}

%% file: main.tex
\section{Abstract}
Model selection for a given target task can be costly, as it may entail extensive annotation of the quality of outputs of different models. 
We introduce \method{}, an efficient method to make an informed decision between candidate text generation models based on preference annotations. \method{} reduces the required amount of annotations, thus saving valuable time and resources in performing evaluation.
\method{} intelligently selects instances by clustering embeddings that represent the semantic differences between model outputs. Thus, it is able to identify a subset of examples that are more informative for preference decisions. Our method is model-agnostic, and can be applied to any text generation model for selecting between models, prompts and configurations. Moreover, we propose a practical iterative approach for dynamically determining how many instances to annotate. In a series of experiments over hundreds of model pairs, we demonstrate that \method{} can dramatically reduce the required number of annotations -- by up to $75\%$ -- while maintaining high evaluation reliability.

\begin{figure}[ht]

    \centering
    \includegraphics[width=0.95\columnwidth]{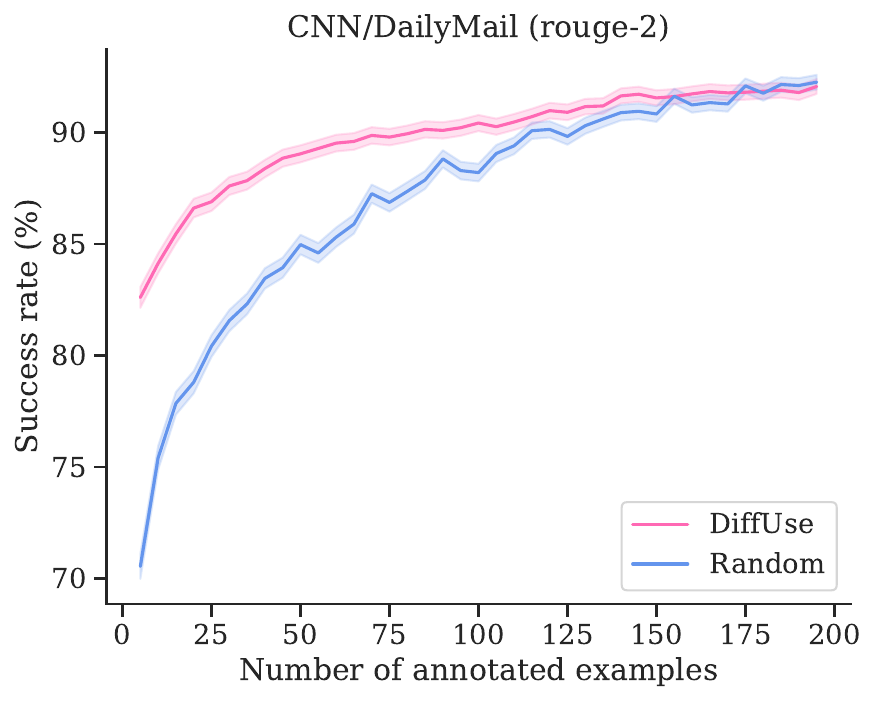} 
    \caption{\textbf{Aggregated success rate} over CNN/DailyMail, 
    across all $666$ model pairs ($\times10$ repetitions for each pair). 
    \method{} demonstrates a clear advantage in correctly determining the stronger model, based on a small number of oracle-annotated examples.
    }
    \label{fig_main_result}
\end{figure}

\section{Introduction}

Model evaluation is a prerequisite for informed decisions -- predominantly, choosing the right model for the task. As such, an essential requirement is the ability to compare models based on how well they perform.

Moreover, since a given model can be configured in many ways, there is a need for an even wider range of comparisons. 
For instance, comparing different prompts a model is provided with can affect task performance significantly.

Comparing model performance generally requires some sort of \textit{oracle} -- a human annotator or LLM-based evaluator 
-- that can judge model outputs and prefer one output over another. However, depending on the nature of the oracle, such judgments can incur significant costs, particularly in terms of annotation budgets \citep{EinDor2019CorpusWA,Lee2019BestPF} and computational requirements \citep{liang2022holistic,biderman2023emergent, perlitz2023efficient}. 
Specifically for text generation tasks, the oracle is burdened with making 
nuanced judgments of the quality of generated texts~\citep{celikyilmaz2020evaluation}; often, this can only be done by expert human annotators~\citep{van2021human}, or possibly by powerful LLMs (e.g., GPT-4, \citealp{zheng2023judging}), both of which are costly to apply at scale. Moreover, as the number of models and tasks increases, conducting these evaluations becomes prohibitively expensive \citep{perlitz2023efficient}.

Our goal is to address the costs associated with evaluating model outputs in text generation, by reducing the burden on the oracle. To our knowledge, this goal has not been addressed in the literature. Specifically, we focus on the use case of directly comparing two candidate models, where the oracle is asked to make preference judgements between the outputs generated by the two models. Our focus is on comparative judgments and not absolute scores, as these are considered more reliable for evaluating text generation~\citep{callison-burch-etal-2007-meta,sedoc-etal-2019-chateval,li2019acute,liang-etal-2020-beyond}.

In this work, we propose a method that substantially reduces the number of examples that must be annotated by the oracle, while yielding a more reliable estimate of the preferred model for the task. 

Our approach - \method{} - selects pairs of model outputs that
on the one hand are representative of the 
space of differences between model behaviors on a given task, 
and on the other hand 
 are more informative, showing clearer preference. Specifically, we calculate embedding vectors that represent the semantic difference between the outputs of the two models; then, by partitioning these embeddings into clusters, we can intelligently select a diverse informative subset of instances for annotation.

\method{} is inherently generic and does not assume anything about the models, tasks, unlabeled test data, prompts, or model hyper-parameters. Our results (\S\ref{sec:results}) demonstrate its stability and effectiveness for different text generation tasks, across hundreds of pairs of generative models, and across a broad range of annotation budgets. One representative example of this can be seen in Figure~\ref{fig_main_result}. We also propose an iterative real-world solution for practitioners (\S\ref{ssec:iterative}), which enables making reliable and cost-efficient choices between candidate models. We find this method to be better in all of our experiments, achieving a reduction in annotations of up to $75\%$ compared to random sampling.

Furthermore, we conduct a comprehensive analysis (\S\ref{sec:analysis}) of the components of our method. Our findings suggest that our method tends to select examples from regions in the output-difference space that are dominated by the preferred model.

\begin{figure*}[htbp]

    \centering
    \includegraphics[width=.8\textwidth]{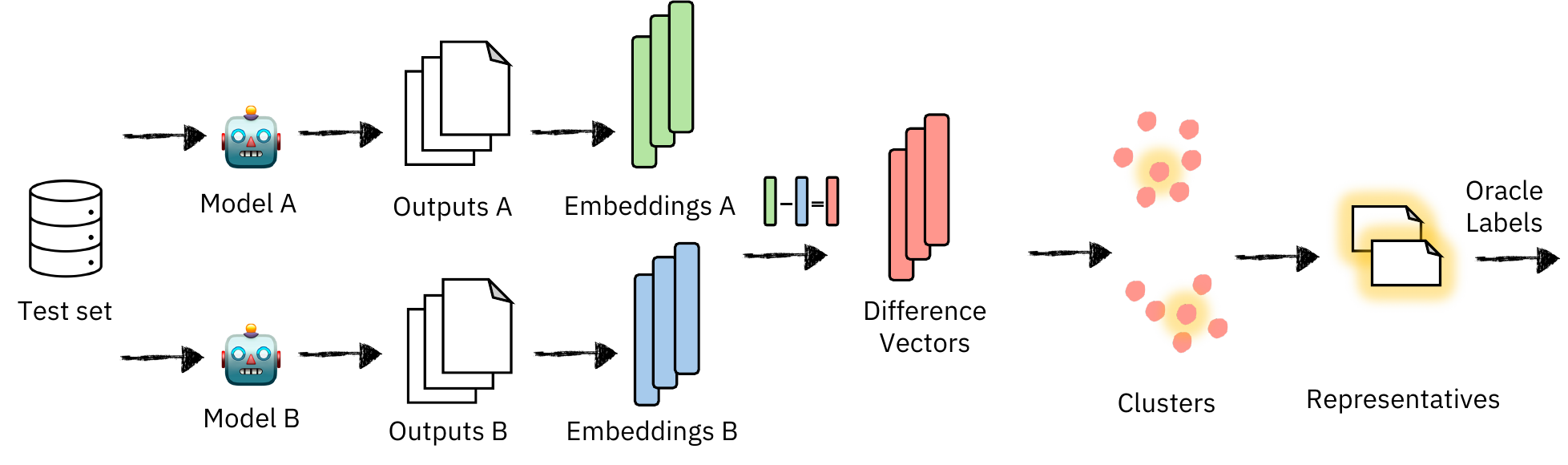} 
    \caption{\textbf{ \method{} flow.} Our method consists of $5$  steps: performing inference with the models on the test set, encoding the generated outputs, performing pairwise subtraction, clustering the resulting vectors, and selecting representatives for evaluation. A comprehensive description is provided in \S\ref{sec:method}.}
    \label{fig_method_flow}
\end{figure*}

\section{Definitions and Problem Formulation} \label{sec:problem_formulation}

In this work, we focus on \textit{comparative} evaluation of models. Given two text generation models, we wish to evaluate which one is stronger with respect to a given generation task, based on preference labels of an oracle over the model outputs.

For an input instance $x$ and a pair of models $M_A$,$M_B$ with corresponding outputs $y_A$,$y_B$, a \textit{preference label} $y_{\text{pref}}\in\{M_A,M_B,T\}$ indicates whether $y_A$ is better than $y_B$ ($M_A$), worse than $y_B$ ($M_B$) or similar to $y_B$ ($T$).  

\paragraph{Test winning model}
The model for which the outputs over the test set $D_{\text{test}}$ are more frequently preferred by the oracle. Formally:
\begin{equation*}
  W_{\text{test}} = 
  \begin{cases}
    M_A & \text{if } P^{M_A}_{\text{test}}>P^{M_B}_{\text{test}}\\
M_B & \text{if } P^{M_A}_{\text{test}}<P^{M_B}_{\text{test}}\\    
T & \text{if } P^{M_A}_{\text{test}}=P^{M_B}_{\text{test}}
  \end{cases}
  \label{eq:w_test}
\end{equation*}
where 
\begin{equation}
P^m_{\text{test}} = \frac{1}{|D_{\text{test}}|}
 \sum_{(x,y_{\text{pref}}) \in D_{\text{test}}} 
 \mathds{1}_{\{y_{\text{pref}}=m\}}
 \label{eq:p_test}
\end{equation}
is the \textbf{test winning probability} of model $m\in\{M_A,M_B\}$, and 
$\mathds{1}_{\{y_{\text{pref}}=m\}}$ is the indicator function that takes the value $1$ if $y_{\text{pref}}=m$ and $0$ otherwise.\footnote{$P^m_{\text{test}}$ is itself an unbiased estimate of $P^m$, i.e., the (unknown) winning probability over all possible input instances from the same distribution.}

\paragraph{Test winning distance}The absolute difference between the 
    test winning probabilities of the two models, $|P^{M_A}_{\text{test}}-P^{M_B}_{\text{test}}|$.
    
\paragraph{Problem formulation}Calculating the test winning model requires preference labels for every point in the test set. However, this is often costly and impractical.
Thus, our goal is to maximize the probability of identifying the test winning model, \textit{under a given annotation budget} $N$, by wisely selecting only a subset of examples
$D_{\text{test}}^{\text{observed}} \subseteq D_{\text{test}}$ from the test set to be labeled by the oracle. 

A naive baseline for estimating the test winning model is to uniformly sample $N$ test instances, label them, and compute the winning model over these instances.

\section{Method} \label{sec:method}

Our algorithm, \method{}, is simple and effective,
and relies solely on the outputs generated by the models. The full flow is described in Figure~\ref{fig_method_flow}.

We aim to represent examples in a manner that captures the distribution of model mismatching behaviors, i.e., various types of differences between model outputs.
To this end, we first embed model outputs into a semantic vector space (using off-the-shelf methods). Subsequently, we generate \textbf{difference vectors} by subtracting the embeddings of one model from the embeddings of the other, for each example in the test set.

Then, we cluster these difference vectors and select one representative from each cluster to be labeled by the oracle. 

Given the construction of the difference vectors, we expect this vector space to largely carry information about semantic differences, i.e., the nature of \emph{disagreements} between models. Choosing an example from each cluster ensures that the set of selected examples is representative of this space; hence, these examples are expected to be informative for estimating which is the preferred model.

\section{Experiments}

\subsection{The Data}

Throughout our experiments, we utilize data from the HELM benchmark \citep[][version 0.2.2\footnote{\url{https://crfm.stanford.edu/helm/v0.2.2/?group=core_scenarios}}]{liang2022holistic}. We rely on data from its core scenarios, which encompass inputs, outputs, and scores for various models, datasets, and tasks.

The 
scores in HELM are automated metrics that compare model outputs to human reference answers, and not direct preference annotations. We chose this data due to its large scale, containing multiple models with their inference results over several well-defined generation tasks. 
The metric scores in HELM serve as the ground-truth data, such that the 
preference label of an example is the model with a higher score for this example (or a tie if scores are equal).

For each scenario we report results for several reference-based metrics, hence simulating a range of different kinds of oracles. 

\begin{figure}[ht]

    \centering
    \includegraphics[width=0.95\columnwidth]{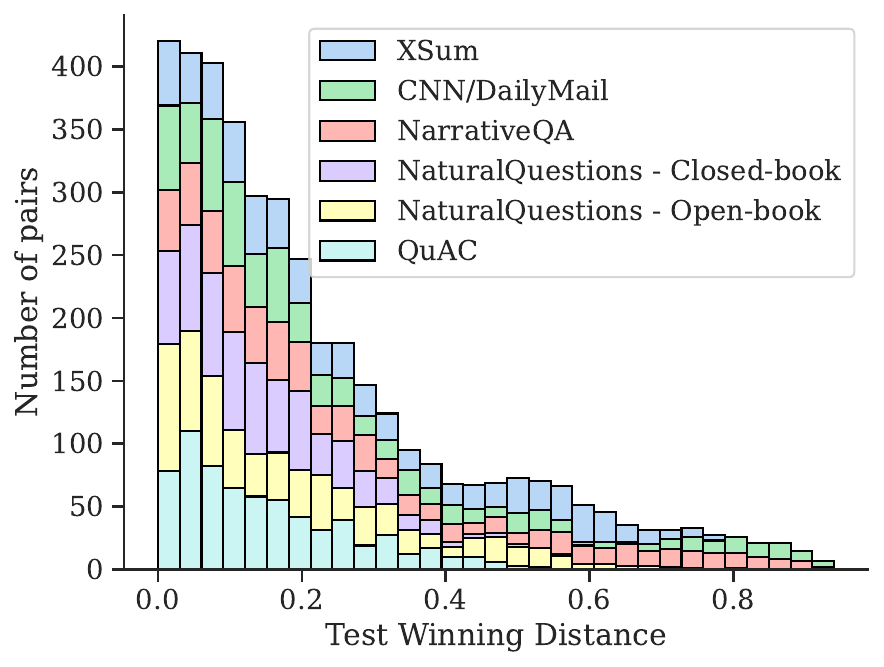} 
    \caption{\textbf{Distribution of test winning distances (\S\ref{sec:problem_formulation}) in HELM} between pairs of generative models.} 
    \label{all_pairs_differences}
\end{figure}

In our experimental setup, we explore $6$ distinct text generation scenarios. This encompasses evaluating the results of $666$ unique model pairings, comprising comparisons among $37$ different models.
Furthermore, we also investigate comparisons involving a single model paired with 3 different versions of prompts (\S\ref{ssec:prompts}), yielding an additional 111 paired comparisons per scenario.
The tasks we experimented with are summarization and question answering (i.e., the text generation tasks in HELM), as detailed in Appendix Table~\ref{table_helm_scenarios}.
As shown in Fig.~\ref{all_pairs_differences}, the winning distances between model pairs in HELM span a large range, but are often small; in other words, HELM showcases diverse behaviors but determining the winning model is usually not trivial.

\begin{figure*}[ht]

  \subfloat{\includegraphics[width=.48\linewidth]{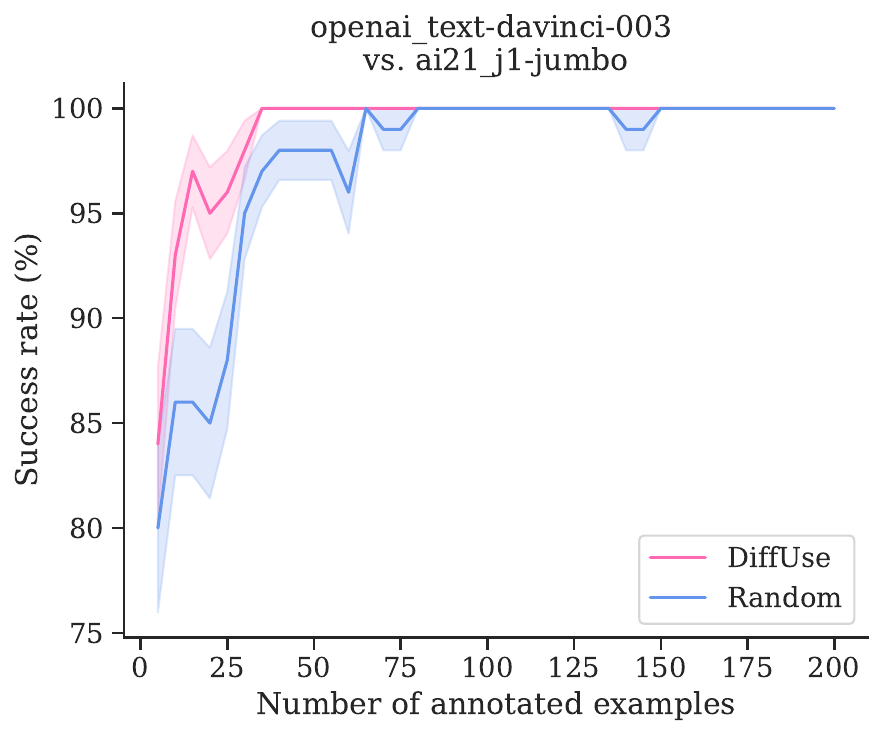}}
  \subfloat{\includegraphics[width=.48\linewidth]{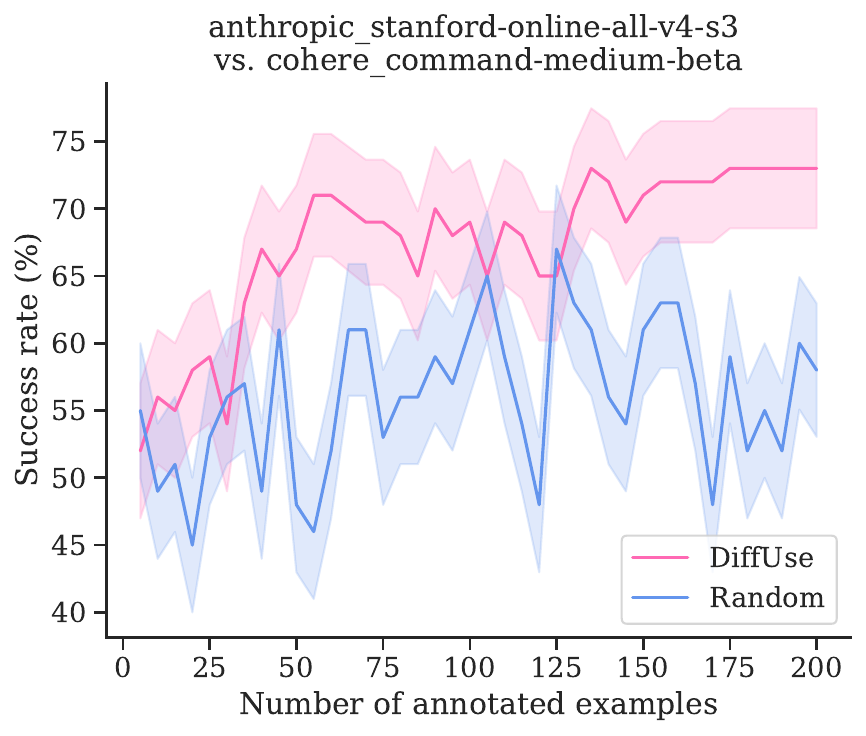}}
    \caption{\textbf{Comparing example selection methods.
    }
    Success rates (± standard error) in identifying 
    the best of two competing generative models (listed in the plot title), in terms of their performance over CNN/DailyMail (using Rouge-2 as the oracle). 
    }
\label{fig_main_result_confidence_example}
\end{figure*}

\subsection{Example Selection Method}

As outlined above (\S\ref{sec:method}), 
\method{} consists of calculating difference vectors that represent the model output behaviors, clustering them, and sampling examples based on the resulting clusters. 

Specifically, we use Sentence-BERT \citep{Reimers2019SentenceBERTSE} \textit{all-MiniLM-L6-v2} encoder to embed the outputs\footnote{\url{https://huggingface.co/sentence-transformers/all-MiniLM-L6-v2}}, and subtract the resulting embeddings to obtain difference vectors. For clustering the vectors, we opt for \textbf{Hierarchical Agglomerative Clustering}~\citep{mullner2011modern} with Ward linkage\footnote{\url{https://docs.scipy.org/doc/scipy/reference/generated/scipy.cluster.hierarchy.linkage.html}} and Euclidean distance.

For a given budget of $N$ examples to be annotated by the oracle, we select them by partitioning the vectors into $N$ clusters. Then, from each cluster we select a single example, and specifically the one whose embedding is closest (in cosine distance) to the center of the cluster.

Note that while we found this setup to work particularly well, opting for a different choice of clustering algorithm, or for a different approach of selecting examples given the clusters, does not dramatically affect the results~(\S\ref{ssec:ablation}).

\subsection{Example Selection Experiments} \label{ssec:selection_exp}

Our main experiments examine the \textbf{success rate} 
of an example selection method, defined as follows.


For a given dataset, a budget of size $N$, and a pair of generative models, we use a selection method to select $N$ examples for annotation. This sample is then annotated by the oracle, and used to determine the \textbf{sample winning model}. An example selection run is successful when the sample winning model equals the test winning model. 
These binary results are then aggregated across several random seeds and across all generative model pairs to determine the success rate of the example selection method. 

\method{} 
is compared to the 
baseline of random selection, where the $N$ examples are sampled i.i.d. from the dataset.

To better estimate the robustness of the selection methods, for each experimental run (seed) we sample a large subset of the full data ($800$ out of $1000$ scenario examples in HELM) and treat this subset as if it were the full test set.

For each of the $6$ HELM scenarios, we report results across $666$ unique model pairs, $10$ runs (seeds) for each, and varying $N$ between $5$ and $200$.

\section{Results} \label{sec:results}

We start by comparing the success rate of \method{} to that of the random selection baseline.

Figure~\ref{fig_main_result_confidence_example} illustrates two such comparisons, 
each for a 
specific pair of models. 
As can be seen, success rates can vary greatly between cases where there is a relatively large performance difference between the generative models (left panel) and those with a small performance difference (right panel). As for the latter, estimating the preferred model is harder and requires more annotated instances. Naturally, the model preference estimation becomes more accurate as the budget $N$ increases and the preference decision relies on a larger set of examples annotated with oracle preference. 

In the two cases presented in
Fig.~\ref{fig_main_result_confidence_example}, \method{} achieves higher success rates at identifying the better generative model, in comparison to random sampling.
These results showcase that with \method{} one can reach the correct decision with a smaller number of examples to be annotated by the oracle.

To give a broader and quantitative picture, 
Figure~\ref{fig_main_result} depicts the aggregated results for the CNN/DailyMail summarization data, averaged across all $666$ model pairs. The plot demonstrates a clear advantage of our approach over random selection, arriving at the correct decision more often and using fewer examples. Thus, using \method{} there is a much lower risk of choosing the wrong model. This pattern is quite consistent across the different datasets tested, as can be seen in Appendix~\ref{app:full_results}. 

Note that while \method{} demonstrates a clear advantage, its effect does vary across datasets, and across ``oracles'' (in our case, different reference-based metrics). 

\input{iterative_algorithm}

\subsection{Estimated Winning Distance} \label{ssec:distance}
Our focus is on making accurate preference choices between models, i.e. choosing the better performing one, according to the oracle preferences. 
However, another facet of model evaluation is the \textit{size} of the performance gap between the two models (e.g., model B won by \textit{18\%} over model A). We define this performance gap, over the entire test set, as the test winning distance (\S\ref{sec:problem_formulation}). 
When using a small set of examples to estimate this performance gap,
we obtain an \textbf{estimated winning distance}. 
Thus, an interesting question is what is the difference between the estimated and test winning distance.  

\begin{figure}[ht]

        \centering
        \includegraphics[width=0.95\columnwidth]{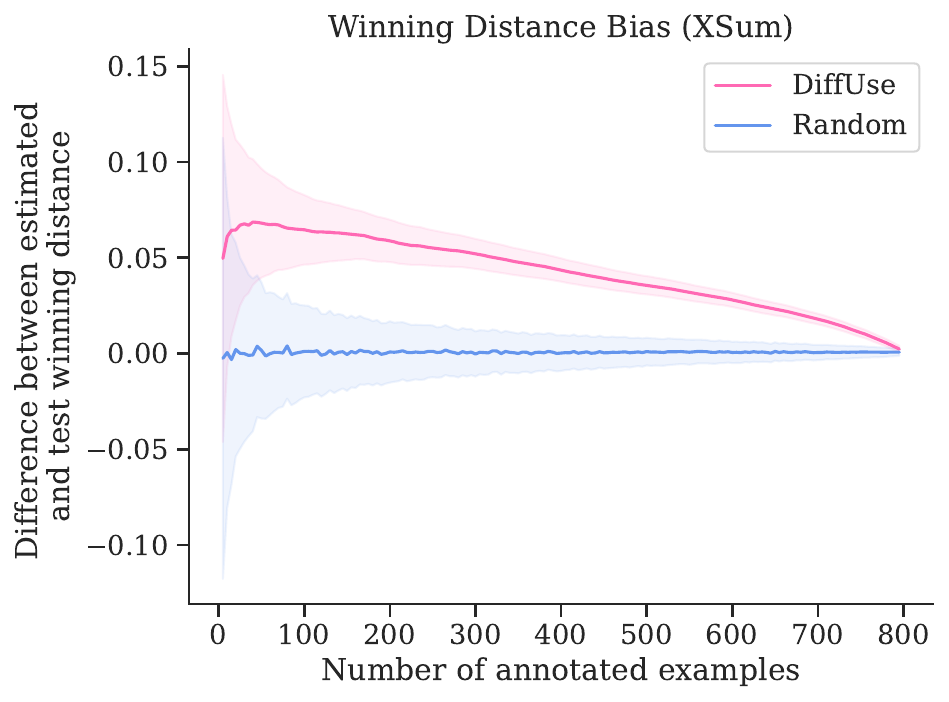}
    \caption{\textbf{Difference between the estimated and test winning distance}, aggregated across all model pairs over XSum. Shaded areas denote standard error (averaged across pairs). Clearly, \method{} favors the test winning model, giving a biased estimate in its favor. The bias dissipates with additional annotations, converging to the true distance for the full set of examples. } 
    \label{fig_diff_distance}
\end{figure}

Figure~\ref{fig_diff_distance} depicts this difference, for each example selection method, over a varying budget size.
Random selection, being an unbiased estimator, naturally has an average deviation of zero from the test winning distance\footnote{This does \textit{not} imply that a single estimation using random selection is likely to be accurate; rather, that across many estimations, 
the \emph{expected value} of 
the difference is zero.}. In contrast, the figure demonstrates that 
\method{} provides an estimated winning distance that is \textit{biased} toward the winning model. This bias, which is particularly large 
with small budgets,
explains how the method is able to outperform random selection at binary preference choices - being biased on average towards the winner, there would also be fewer cases where the losing model is accidentally
selected (note the lower bounds of the shaded areas in Fig.~\ref{fig_diff_distance}).

\subsection{Practical Iterative Selection Algorithm} \label{ssec:iterative}
Accuracy in estimating the winning model can vary widely, depending on the budget size as well as the actual performance gap (Fig.~\ref{fig_main_result_confidence_example}). 
In a 
real-world 
scenario, however, users 
do not know in advance the size of the performance gap between the models they compare; moreover, after annotating some examples with the oracle and estimating the winning model, users \textit{will not know whether the estimation is in fact correct}.

Thus, in order to 
reduce oracle effort 
in practice, there is a need for an
approach that determines the 
minimal budget required, 
and provides some approximation of the reliability of the winning model selection.
To this end, we propose an \textbf{iterative method} for selecting examples. In this approach, the number of examples sent to the oracle is increased gradually, until a predefined reliability-oriented threshold is met. Hierarchical clustering naturally lends itself to an iterative solution: suppose we have clustered the difference vectors into $k$ clusters, and the oracle has annotated the $k$ selected examples, yet we suspect that the preference estimation is not sufficiently reliable. In this case, we can now cluster the vectors into $k+1$ clusters; this will further partition one of the previous clusters, providing two new examples to be labeled by the oracle\footnote{Partitioning a cluster means selecting two new examples, in addition to the one originally annotated for the cluster; we discard the original example ($e_c$ in Alg.~\ref{algo:model_comparison}) from the preference decision, as it is presumed to be less informative at this point.}. With each partitioning step, the amount of information increases, and this procedure is repeated until reaching the threshold/stopping criterion.

The full iterative selection flow is described in Algorithm~\ref{algo:model_comparison}. For the stopping criterion, we propose a reliability threshold based on the hypergeometric distribution (for details, see App.~\ref{app:hypergeometric}). The threshold 
is a heuristic that approximates the level of risk, where a 
threshold of $0.1$, for example, loosely corresponds to a likelihood of up to $10\%$ of choosing the wrong model. The 
threshold is set in advance, and reflects a preferred point on a trade-off: between the user's tolerance for error, and the amount of examples the oracle will need to annotate.

Results for the iterative algorithm are shown in Figure~\ref{fig_iterative}. Clearly, 
\method{} provides a significant advantage over random selection, increasing the likelihood of successfully determining the winner (right panel), while significantly reducing the number of examples sent to the oracle (left panel).

Note that the number of annotations in practice varies widely, and is linked to the performance gap between the models. For instance, in the Closed-Book version of NaturalQuestions, a large number of examples is annotated, and the outcome is usually inconclusive (left panel of Fig.~\ref{fig_iterative}, App. Tab.~\ref{table_iterative}); 
the reason for this is that
the test winning distances  
in this dataset are quite small 
(cf. Fig.~\ref{all_pairs_differences}), making it difficult to conclusively determine the winner.

\subsection{Prompt Selection} \label{ssec:prompts}
Naturally, task performance varies depending on the underlying model used. However, there are additional configurations affecting downstream task performance. One such crucial aspect is the choice of prompt and of in-context examples~\cite{polo2024efficient, mizrahi2023state}.

Thus, we also test our approach in distinguishing between different prompts and in-context exemplars. Specifically, for each model and scenario, we apply our method to the model outputs using different prompt variants. As done with outputs from different models, the instances selected with \method{} are then used to estimate the the better performing variant.
We utilized the scenario data provided by HELM, which includes three prompts for each model with variations on 
the few-shot exemplars given before the input.

We find that akin to the between-model experiments, 
our method is also effective in identifying 
the better prompt for a given model, using
much fewer samples than a random selection. The results, depicted in Appendix~\ref{app:prompt}, are
consistent across datasets, tasks and scores.

\input{plots/fig_iterative}

\section{Analysis} \label{sec:analysis}
\subsection{Method Parameters} \label{ssec:ablation}
Next, we examine $3$ components
in the flow of \method{} (Fig.~\ref{fig_method_flow}): the \textbf{representations} of examples,
the \textbf{clustering algorithm} and the \textbf{cluster representative selection}.

Our method relies on difference vectors (i.e., subtraction of output embeddings) to represent examples. A naive alternative 
would be to cluster the embeddings of \textit{inputs}, akin to some methods in active learning~\citep{zhang-etal-2022-survey}. However, we find that this approach does not consistently outperform random sampling (App. Figure~\ref{fig_inputs_baseline}).
\cready{We also compare various approaches for aggregating the semantic embeddings of the two \textit{outputs}. We find that using difference vectors, i.e., subtracting the two embeddings, outperfoms other aggregation methods, such as concatenation or addition (App. Figure~\ref{fig_}). \ag{add}}

In contrast, we find that the choices of clustering algorithm and representative selection are less significant, and performance differences are not dramatic (Appendix~\ref{app:clustering_methods}). Note that all configurations significantly outperform the random baseline.

\subsection{Which examples are selected?} \label{ssec:norm_analysis}

As shown above, the success of our method hinges on the use of output difference vectors. Next, we perform several analyses to better understand how clustering these vectors enables selecting examples that are informative for the oracle.

The difference vectors represent variance in the outputs, and thus in the models' behavior for a given task. Assuming an ideal semantic encoder, highly distinct outputs should yield difference vectors with high norms, signifying pronounced dissimilarities. Conversely, similar outputs would result in lower norms, indicating subtle differences.

\begin{figure}[htbp]

    \centering
    \includegraphics[width=0.95\columnwidth]{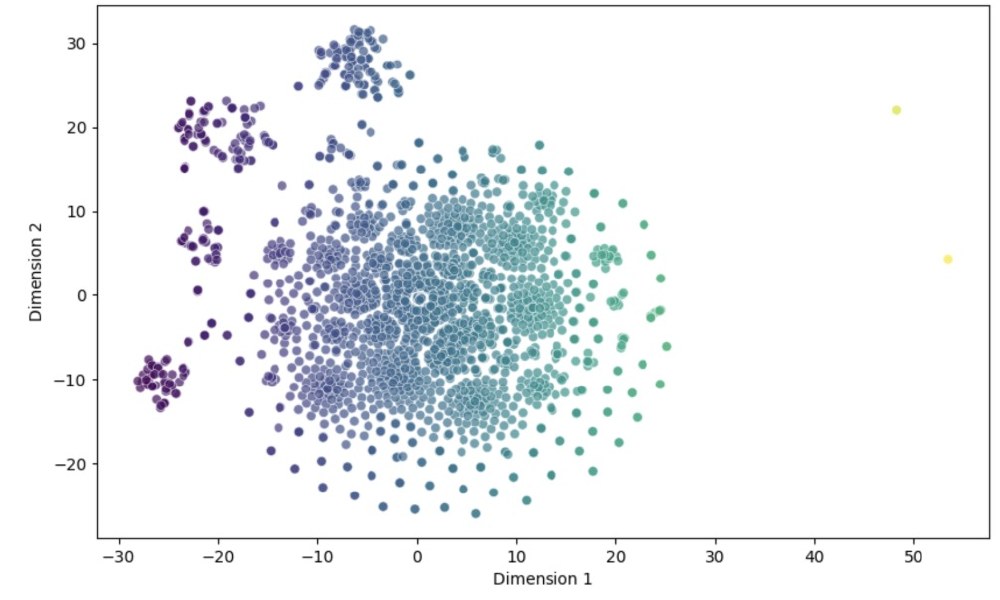} 
    \caption{\textbf{Example 2-D projection.} A t-SNE \citep{tSNE} projection of the difference vectors from a randomly selected pair of models in XSum. The observed behavior, where
    most vectors are centered around zero, and the distribution is sparser away from it, is consistent across model pairs.}
    \label{tsne_diff_vecs}
\end{figure}

\subsubsection{Cluster Sizes and Difference Norms} \label{subsection_clusters_size_norm}

In distance-based clustering,
vectors with smaller norms have a higher tendency to be clustered together.
This is nicely demonstrated in Figure~\ref{tsne_diff_vecs}, which depicts an example two-dimensional projection of difference vectors for a pair of models. The projection reveals 
a densely populated region close to zero, corresponding to cases where the model outputs show more subtle differences.

\begin{figure}[!ht]

    \centering
    \includegraphics[width=0.92\columnwidth]{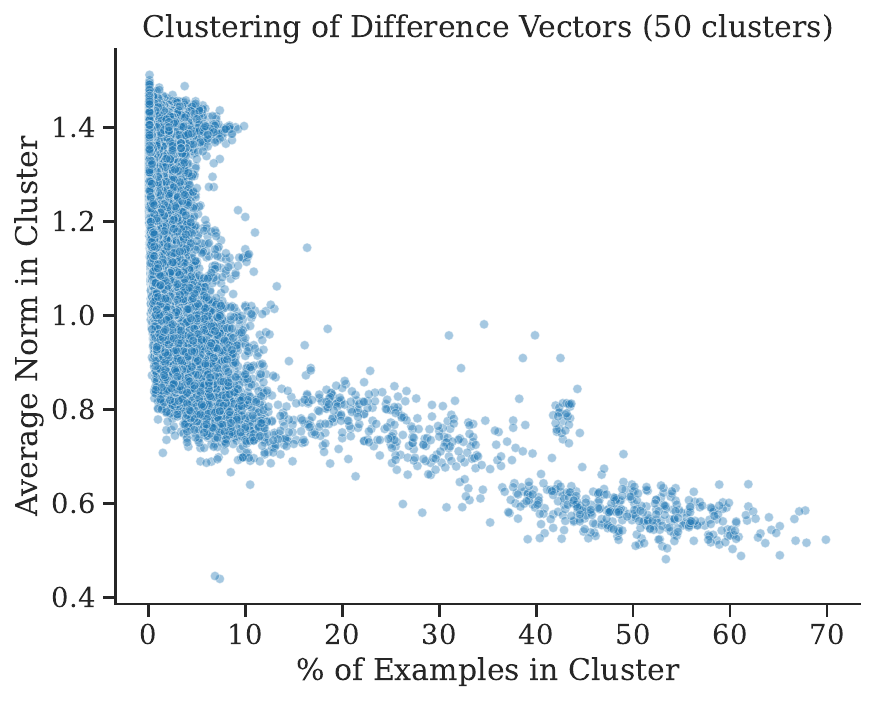} 
    \caption{\textbf{Cluster size vs. average vector norm.} 
    Hierarchical clustering results of the difference vectors, partitioning XSum into $50$ clusters.
    Each point represents a single cluster; in total, the plot depicts $\sim33$K points ($666$ model pairs  $\times~50$ clusters per pair).
    The x-axis reflects the percentage of all examples that are in the cluster (i.e., indication of cluster size), and the y-axis is the average vector norm within the cluster. Results are characterized by a few very large clusters with a small average norm (bottom right); this pattern is consistent across different numbers of clusters (App. Fig.~\ref{fig_cluster_size_norm}).}
    \label{clusters_rel_size_norm}
\end{figure}

Figure~\ref{clusters_rel_size_norm} illustrates the relation between the sizes of clusters and the average norm of difference vectors within the cluster. Evidently, clustering the difference vectors tends to result in a small number of large clusters, which have a low average norm (bottom-right area of Fig.~\ref{clusters_rel_size_norm}), alongside a large number of small clusters with higher norm values. Often, over half of the vectors are assigned to a single cluster with small norms.
 As \method{} selects one example from each cluster, the sub-population of examples with small difference norms is under-represented in the set of selected examples.

\begin{figure}[ht]

    \centering
    \includegraphics[width=0.98\columnwidth]{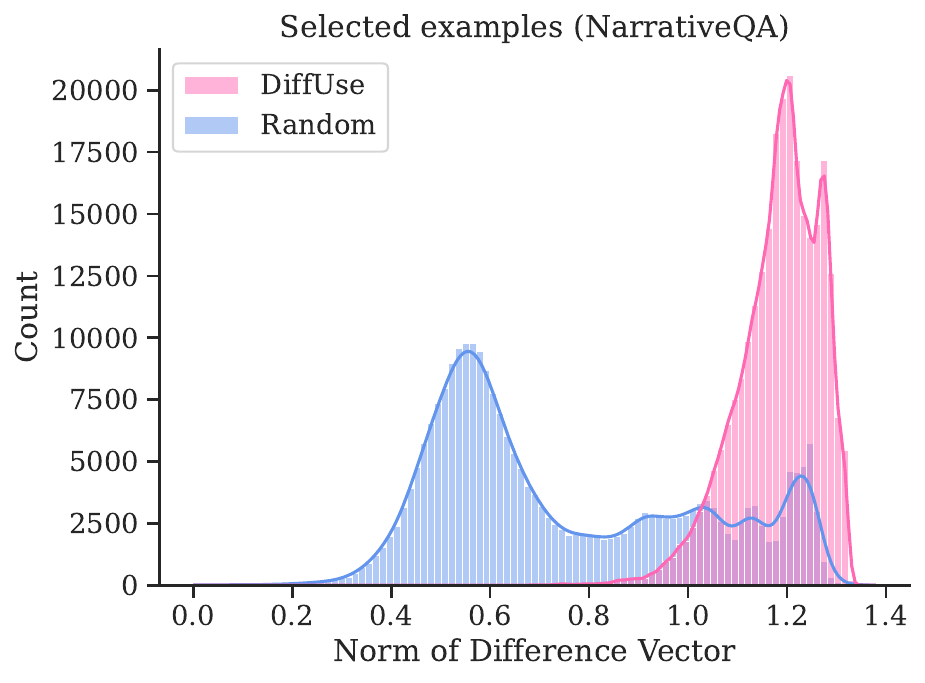} 
    \caption{\textbf{Norms of selected examples.} The histograms depict the norm of difference vectors for the output pairs selected for annotation (across all NarrativeQA selection runs). 
    Compared to random sampling,
    \method{} selects examples with higher vector norms.}
    \label{corr_norm_selections}
\end{figure}

Figure~\ref{corr_norm_selections} directly depicts the norm size distribution \emph{of the selected examples}. Again, we see that \method{} is biased toward high-norm instances.

\subsubsection{Norms and Winning Model} \label{subsection_winning_norms}

We have demonstrated that our method over-represents difference vectors with a higher norm. This leads to the question of how this tendency relates to model preference. 

\begin{figure}[ht]

    \centering
    \includegraphics[width=0.93\columnwidth]{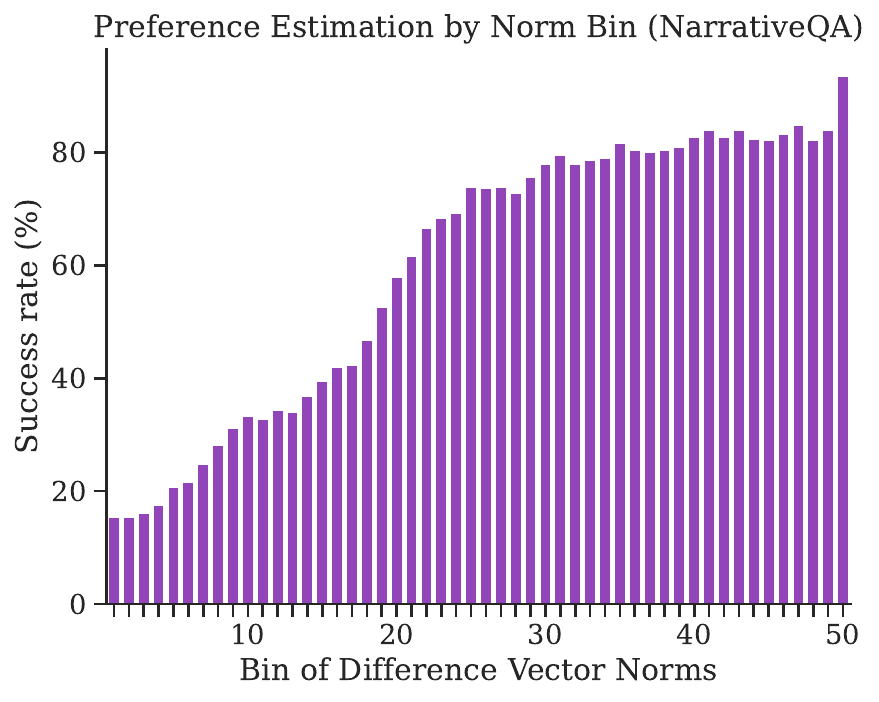} 
    \caption{\textbf{Norms and preference estimation}. The plot depicts the success rate at estimating the test winning model, based on sub-populations with varying vector norms. For each model pair, the difference vectors were partitioned based on their norm sizes into $50$ equal-count bins, and each bin of instances was used to estimate the test winning model. The plot presents an aggregation across all model pairs over NarrativeQA.}
    \label{corr_norm_wins}
\end{figure}

Figure~\ref{corr_norm_wins} depicts the relation between the norm of difference vectors and estimation of the test winning model. As can be seen, the preference label of instances with higher difference norms is more likely to align with the test winning model. This is in line with the winner-bias shown in Fig.~\ref{fig_diff_distance}.

A possible explanation for this observation is that
larger semantic differences between the models' outputs are expected to be associated with larger quality gaps; meanwhile, the chances that the weaker model will beat the stronger model's output by a large margin are low. Thus, the lower the difference norm, the higher the probability of the preference label to be ``erroneous'', namely for the weaker model to be preferred by the oracle. 


Given that high-norm pairs are informative, a simple approach would be to forgo clustering, and simply select the instances with the highest norm for annotation. However, this results in inferior performance (App.~\ref{app:norm}), likely due to low diversity and representativeness of the selected subset. 
This is not surprising; selecting by norm alone can result in outliers, and may not be representative of the space of difference vectors.


To sum, clustering difference vectors over-represents output pairs with a large difference norm (\S\ref{subsection_clusters_size_norm}). 
These, in turn, are more strongly associated with the winner (\S\ref{subsection_winning_norms}). 
Thus, our analyses illustrate how \method{} is able to correctly determine the test winning model using fewer annotations.

\section{Related Work}
In light of the soaring costs of language model evaluation, even when using automatic metrics, some recent works \citep{perlitz2023efficient, maynez-etal-2023-benchmarking} have studied the effects of reducing the size of evaluation sets -- via random sampling -- on the reliable ranking of models. 

Other prior works have examined methods of \emph{intelligently} selecting subsets of examples for evaluation, aiming to find sets of examples that are more informative than randomly sampled instances. 

\citet{rodriguez-etal-2021-evaluation,vania2021comparing} look at 
selecting examples for evaluating new models, given fully-annotated question answering data for an existing set of models. 
They show that some selection strategies, based on item response theory \citep{lord1968statistical}, outperform the random selection baseline for ranking new models on a question answering task.
Several works have addressed 
label-efficient assessment in the context of \textit{classifier} performance. 
\citet{katariya2012active} propose a label-efficient algorithm to gain better accuracy estimates of classifiers, by selecting examples to label based on stratified sampling.  \citet{Ji_Logan_Smyth_Steyvers_2021} suggest an active Bayesian approach that uses inferred uncertainty to guide selection of instances. 
Inspired by works on active learning, 
\citet{kossen2021active} propose methods based on a stochastic acquisition process, to avoid unwanted and highly problematic biases involved in active selection of test set examples. 
\citet{ha2021alt} suggest an iterative method that utilizes a surrogate model to estimate the metrics of interest over the unlabeled test set, and labels examples that lead to maximal uncertainty reduction of the metric estimation. With a similar spirit to our work, \citet{vivek2023anchor} find anchor examples in classification datasets that represent how confident different models are over those input examples.

Our work differs from these prior efforts in that we tailor our approach to the nature of text generation. Existing methods for example selection are not easily adapted from classification to generation tasks -- the concepts of uncertainty, confidence and errors are inherently different for natural language generation, necessitating a different approach. In addition, unlike e.g., \citet{rodriguez-etal-2021-evaluation,vania2021comparing}, our method does not require any annotations and assumes only a set of model outputs.

\cready{The current work also draws inspiration from works finding patterns in model behaviour, for instance that models seem to learn in a similar order and make the same mistakes \citep{choshen2022grammar,hacohen2020let}, and that evaluation trained on one model can work well over another \citep{wan2022unite} or even work on the input level alone \citep{don-yehiya-etal-2022-prequel}. 
More closely, our work relates to efforts aimed at reducing annotation costs during training, namely iterative active learning approaches \citep{zhang-etal-2022-survey,ein-dor-etal-2020-active,perlitz-etal-2023-active}.}

\section{Discussion}
We have demonstrated that our method, \method{}, provides significant cost savings in model selection. 
We tested the approach for choosing the better underlying model as well as the most effective in-context prompt. Given the generality of the method, this likely means \method{} is applicable to a vast range of model configuration and hyper-parameter choices.

Moreover, using a dynamic algorithm such as the one proposed here (\S\ref{ssec:iterative}), practitioners can 
reduce the 
number of oracle judgements while maintaining high evaluation reliability.

\cready{In this work we provide a somewhat novel view into the notion of bias. Bias in an example selection method is often seen as a problem, one that should be avoided or corrected~\citep{kossen2021active}. In contrast, in our setting the power of \method{} is precisely the fact that it is biased in a very particular direction (Fig.~\ref{fig_diff_distance}, \S\ref{ssec:norm_analysis}), reducing the likelihood for error in the context of binary preference decisions.}

\cready{The proposed approach can be easily adapted to domains beyond NLP such as vision and speech.}
Here we examined the problem of selecting between a pair of candidate models. We leave to future work the scenario of picking from a larger set of candidates. This may entail adapting our method to a multi-model scenario, or combining our pairwise approach with an efficient method for limiting the number of pairwise comparisons (e.g.,~\citealp{mohankumar-khapra-2022-active}).

While the current work deals with model selection, our approach of modeling differences between outputs can potentially be applicable for other purposes as well. This can include qualitative assessment of model behaviours, collection of preference data for training reward models, and more. Moreover, the proposed approach can be easily adapted to domains beyond NLP such as vision and speech.

\section*{Limitations}
As our approach relies on obtaining representations of model outputs, it incurs the non-trivial computational cost of performing inference over the set of examples to be clustered, in the range of hundreds of examples. Thus, our method is only suited for the (very common) scenario where the cost of applying the oracle is significantly greater than the cost of performing inference on a somewhat larger set of examples. This is the case for example when the oracle is a paid API or a human annotator.

As noted in \S\ref{ssec:distance}, \method{} is a biased approach that tends to over-represent subpopulations of the of examples. Here we show empirically -- across model pairs and across datasets -- that this method provides significant and consistent gains in relation to random selection. However, as also mentioned in App.~\ref{app:hypergeometric}, for a given attempt at model comparison there is no theoretical or statistical guarantee of the probability of making the correct choice.

Our study is motivated by the fact that obtaining a large amount of quality or preference judgments for a target generation task and candidate models is prohibitively expensive. Ironically, this also means it is not trivial to obtain large-scale  annotated data that can be used for \emph{evaluating} the accuracy of our oracle minimization approach (existing multi-model datasets, e.g. for RLHF, often do not have a well-defined notion of target tasks). Hence, here we rely on reference-based metrics in HELM to simulate different types of oracles. This is a limitation of this work as we do not directly demonstrate our method on real-world preference oracles.

%% file: iterative_algorithm.tex
\begin{algorithm*}[t]

\caption{Iterative Selection Algorithm - Risk-based Threshold}
\label{algo:model_comparison}

\textbf{Input:} Two models $\{M_A, M_B\}$, dataset $D$, and oracle $O$

\textbf{Parameters:} Threshold $p \in (0, 1)$, Minimum number of annotations $n$, Maximum budget $N$

\textbf{Output:} Winning model (or inconclusive)

Calculate the difference vectors $V(D, M_A, M_B)$, as described in Section \ref{sec:method}.

Cluster $V$ into $n$ clusters.

Choose representatives $E_n = \{e_1, \ldots, e_n\}$, one from each cluster.

Get the oracle tags $T_n$ = $O(E_n)$, and calculate the probability $sf_{\text{hypergeom}}(T_n)$ (see App.~\ref{app:hypergeometric})

Initialize $k = n+1$, $T = T_n$

\While {$sf_{\text{hypergeom}}(T) > p$ and 
$|\text{labeled examples}| < N$
}{

    Find the next cluster $c$ to be split $(1 \leq c < k)$, and split it.
    
    Choose representatives $E_k = \{e_k, e_{k+1}\}$, one from each of two splits.
    
    Get the oracle tags $T_k$ = $O(E_k)$.

    $k = k+1$, $T = (T_n - \{e_c\}) \cup T_k$

}

Return the winning model according to $T$.
\end{algorithm*}

%% file: plots/fig_iterative.tex
\begin{figure*}

    \subfloat{\includegraphics[width=0.48\textwidth,valign=t]{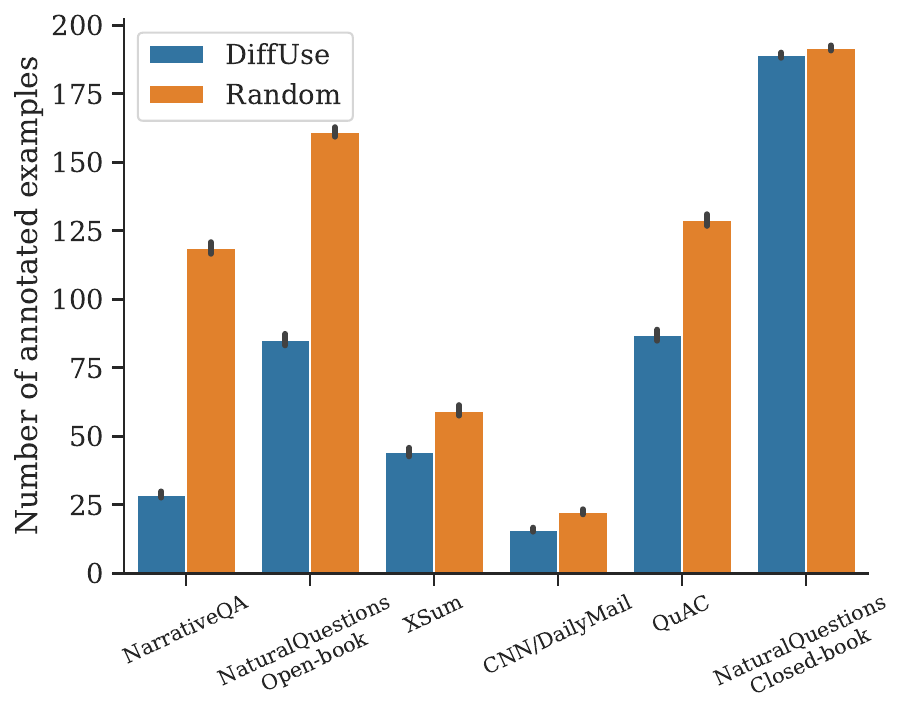}}
    \subfloat{\includegraphics[width=0.48\textwidth,valign=t]{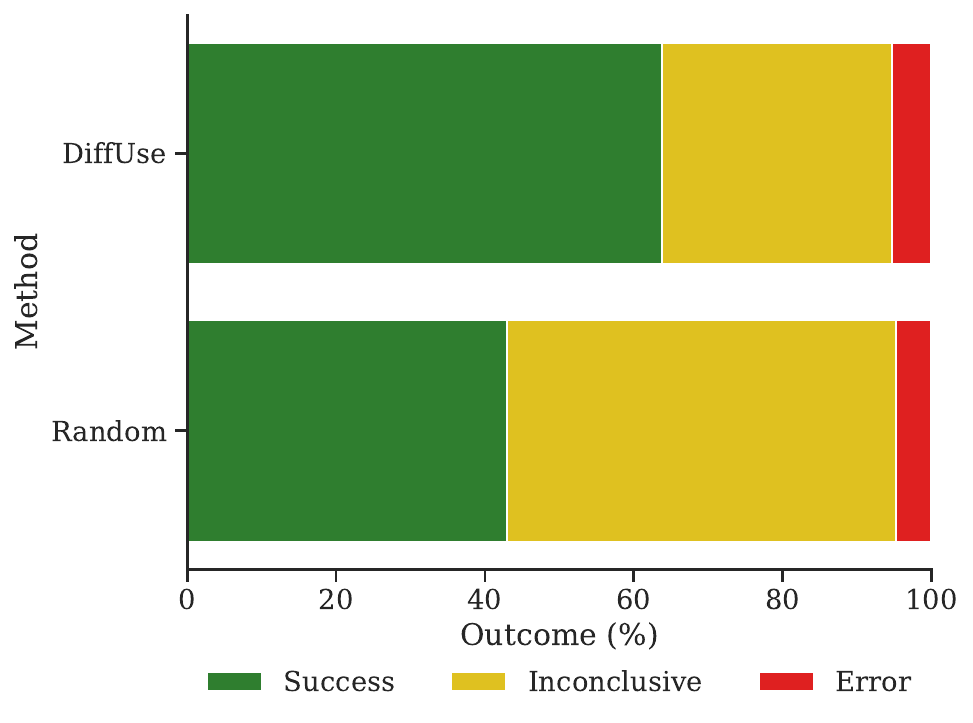}}

\caption{\textbf{Iterative selection results}
(Algorithm~\ref{algo:model_comparison}; with $p=0.2$, $n=5$, and $N=200$), comparing \method{} to random sampling. Results are aggregated across $666$ model pairs. The left panel depicts the mean number of examples annotated by the oracle before reaching the stopping criterion. The right panel depicts the proportion of outcomes of the iterative selection experiments -- i.e., was a winning model determined, and was this decision correct -- aggregated across all datasets. See also App.~Table~\ref{table_iterative}, \ref{table_iterative_0.1}.}
\label{fig_iterative}

\end{figure*}

%% file: appendix.tex
\newpage
\section{Appendix}

\input{datasets_table}

\subsection{Full Results} \label{app:full_results}
Results for the $6$ text generation scenarios (datasets) in HELM, with $3$ different metrics for each scenario, are presented in Figure~\ref{fig_full_results}.

\subsection{Computational Budget}
Our example selection results (e.g., in Figures~11-14) consist of $\sim 1.6$ million selection runs, for every selection method: $6$ scenarios $\times$ $666$ model pairs $\times$ $40$ annotation budgets (between $5-200$) $\times$ $10$ repetitions (seeds).

The HELM raw data already includes the model inference outputs as well as the preference judgements (metric scores) of the different models. Thus, the computational costs of performing these experiments consist mainly of the semantic encoding of the model outputs, as well as clustering of the representation vectors. The semantic encoding, using S-BERT ($1000$ examples per scenario $\times$ $37$ models in HELM) took a few minutes per scenario on a single GPU; most of the computational cost consisted of a large number of clustering runs, which were performed in parallel on $16$ CPU cores.

\input{plots/app_fig_all}

\subsection{Iterative Selection Threshold} \label{app:hypergeometric}

As described in \S\ref{ssec:iterative} and Algorithm~\ref{algo:model_comparison}, we propose an iterative algorithm for annotating examples by the oracle and choosing the winning model.

We opt for a reliability-oriented stopping criterion that is based on the hypergeometric distribution. This distribution describes the probability of `success' when sampling without replacement, and is parameterized by a population size $N$, sample size $n$, number of successes in the population $K$ and number of successes in the sample $k$.

Specifically, we look at the hypergeometric distribution survival function, $sf_{hypergeom}(k-1)$, which describes the probability of getting $k$ or more successes by chance. In a model comparison scenario, $n$ corresponds to the number of examples annotated by the oracle, and $k$ to the number of votes received by the winning model within this set. We define the null hypothesis as one where the winning model is the winner in $50\%$ of the instances in the full test set, i.e., where $K = N/2$. Using this value for $K$, The result $sf(k-1)$ thus reflects how likely or unlikely it is to get a value of $k$ or higher \textit{given a ground-truth $50\%$ win rate}.

For instance, say we select examples out of a pool of $500$ unlabeled examples. The oracle is given a total of $10$ examples to label, and determines that model $A$ was the winner in $8$ of them:
\begin{equation}
    sf(k-1,N,K,n) = sf(7,500,250,10) = 0.0529 \nonumber
\end{equation}

Thus, in this example -- given the null hypothesis and assuming a hypergeometric distribution -- there is only a $\approx{5\%}$ probability of getting such a high win rate -- or a higher one -- by chance. In other words, a situation where  model $A$ is the winner in just $50\%$ of the full test set, and an $8/10$ result was obtained, is relatively unlikely. A situation where model $A$ is the winner in \textit{under $50\%$} of the test set is even \textit{less} likely. This means that the user can be fairly confident that the correct winner was chosen.

Thus, when applying the iterative algorithm, the user sets an acceptable risk level -- say, 10\% -- in advance; at each iteration, $sf$ is calculated using the current values of $n$ and $k$; if the value of $sf$ is lower than the risk level, the result is considered sufficiently reliable; if not, the sample size $n$ is increased and additional examples are labeled.

Note that we use this probability-based threshold merely as a heuristic, or proxy, for the real probability. In practice, the assumptions of the hypergeometric distribution are violated in our case. Most importantly, this distribution describes random selection, whereas \method{} is non-random, and in fact has a distinct bias towards selecting certain kinds of examples (\S\ref{ssec:distance}, \S\ref{sec:analysis}). Moreover, even for random selection, the approach does not precisely match the model comparison setting; for instance, if there is a large number of examples where there is a tie between the two models, a null hypothesis of a $50\%$ win-rate is in fact overly conservative. Thus, while the threshold chosen by the user serves as a good proxy for the estimated error rate, and is thus suitable as a stopping criterion, it does not guarantee the actual error rate value. In our empirical experiments, for all datasets the error rate was lower than the chosen risk threshold (cf. Tables~\ref{table_iterative},\ref{table_iterative_0.1}).

When opting for higher risk thresholds, there is a large impact to the initial number of labeled examples,
because wins that are based on a very small sample (e.g., $3$ out of $3$) are avoided, even though they may meet the risk threshold.

\input{iterative_table}

\subsection{Clustering Methods and Representative Selection} \label{app:clustering_methods}



We conducted 
selection experiments
employing various clustering algorithms. We found that the majority of these algorithms produced results that exceeded those of random sampling.

Below, we provide details regarding the clustering methods we explored:

\begin{enumerate}
\item \textbf{Hierarchical Clustering}
\begin{enumerate}
\item \textbf{Euclidean Distance:} Hierarchical clustering with Euclidean distance measures dissimilarity between data points based on their spatial coordinates. It facilitates cluster creation by iteratively merging data points to minimize within-cluster variance.
\item \textbf{Cosine Distance:} Hierarchical clustering using cosine distance measures similarity between data points via the cosine of the angle between vectors. Cosine distances were employed during the merging process.
\end{enumerate}
\item \textbf{K-Means Clustering:} K-Means clustering partitions data into 'k' clusters by iteratively assigning data points to the nearest cluster center and updating centers based on the mean of assigned points. Our approach incorporated ``greedy k-means++'' for centroid initialization, leveraging an empirical probability distribution of points' contributions to overall inertia.
\end{enumerate}

The model preference success rates for different clustering algorithms, selecting a single representative from each cluster based on distance to the cluster center, are shown in Figure~\ref{fig_clustering}.

We also explored various methods for selecting a \emph{representative} from each cluster. These methods encompassed random selection, choosing the example nearest to the centroid (employing either Euclidean or cosine distances), and selecting the example with the maximum norm. As seen in Figure~\ref{fig_representatives},  the choice of representatives did not significantly impact the outcomes.

Here we focus on clustering algorithms as the approach for sampling from the vector distribution. However, other selection approaches, such as core-set \citep{sener2018active} or IDDS \citep{tsvigun-etal-2022-active}, may also prove effective.

\input{plots/app_fig_clustering}
\input{plots/app_fig_representatives}

\input{plots/app_fig_inputs}


\begin{figure*}[h]
    \subfloat{\includegraphics[width=.33\textwidth]{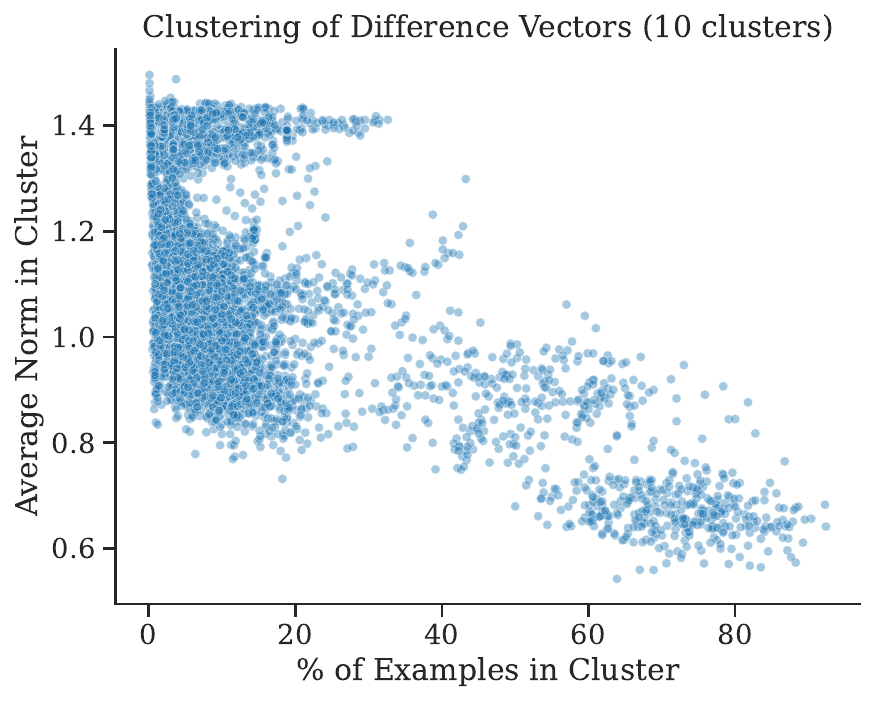}}
    \subfloat{\includegraphics[width=.33\textwidth]{final_plots/analysis/cluster_size_norm_50.pdf}}
    \subfloat{\includegraphics[width=.33\textwidth]{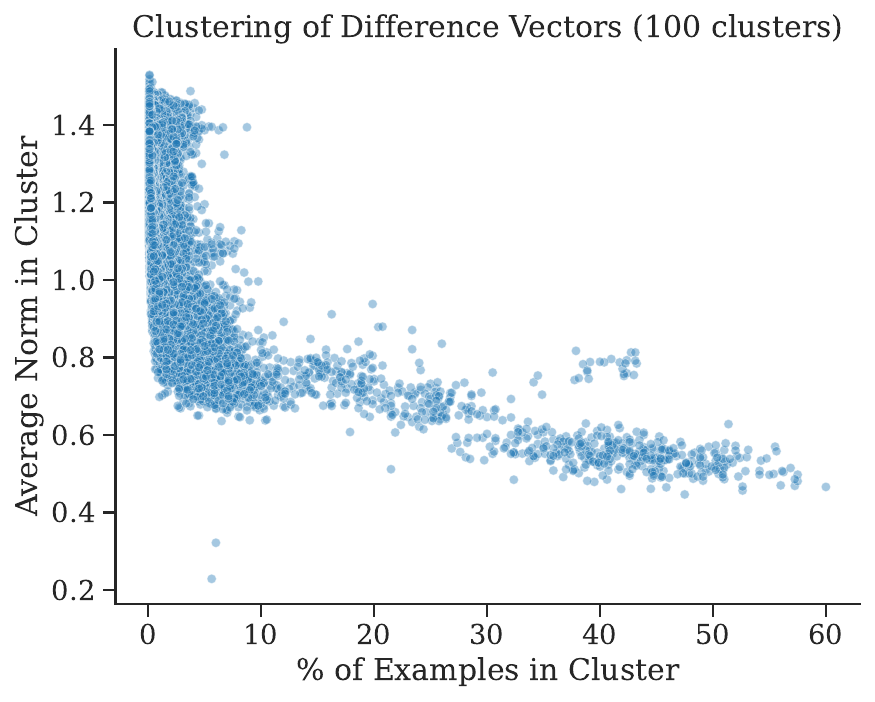}}
  \caption{\textbf{Cluster size vs. average vector norm.} 
    The plot describes the results of hierarchical clustering of the difference vectors, for the XSum dataset when partitioning into different numbers of clusters.
    Each point represents a single cluster; in total, each panel depicts between $6.7$K and $67$K points ($666$ model pairs  $\times$ the number of  clusters per pair).
    The x-axis reflects the percentage of all examples that are in the cluster, and the y-axis is the average vector norm within the cluster. The results are characterized by very large clusters with a small average norm (bottom right of the plots).} \label{fig_cluster_size_norm}
\end{figure*}

\subsection{Norm of Difference Vectors} \label{app:norm}

We explored the norm of the difference vectors as a signal for selecting examples. While we experimented with various binning scenarios, the best outcomes were obtained by directly selecting the vectors with the maximal norm. However, even this approach proved inconsistent across datasets and tasks, as demonstrated in Fig.~\ref{graphs_norm_baseline}. This is not surprising; selecting by norm alone can result in outliers, and may not be representative of the space of difference vectors.

\subsection{Prompt Selection Results} \label{app:prompt}

Results for the prompt choice experiments are presented in Figure \ref{fig_prompts}. The results span the $6$ text generation scenarios (datasets)
in HELM, with $3$ different metrics per scenario. The plots in Figure \ref{fig_prompts} aggregate multiple paired selection experiments, where in each experiment the choice is between two prompt variants used with the same underlying model. For each scenario, and for each of the $37$ models in HELM, $3$ prompt variants were tested; thus, each panel depicts $111$ unique paired comparisons.


\input{plots/app_fig_max_norm}





\input{plots/app_fig_prompts}

%% file: datasets_table.tex
\begin{table*}[htbp] 
    \begin{center}
    \begin{tabular}{|p{3cm}|p{2.5cm}|p{9cm}|}
    \hline
    Task & Scenario & Description \\
    \hline
    \multirow{2}{*}{Question Answering}
    & NarrativeQA & The NarrativeQA benchmark for reading comprehension over narratives \cite{kocisky-etal-2018-narrativeqa}\\
    & NaturalQuestions (closed-book) & The NaturalQuestions \cite{kwiatkowski-etal-2019-natural} benchmark for question answering based on naturally-occurring queries through Google Search. The input does not include the Wikipedia page with the answer. \\ 
    & NaturalQuestions (open-book) & The NaturalQuestions \cite{kwiatkowski-etal-2019-natural} benchmark for question answering based on naturally-occurring queries through Google Search. The input includes the Wikipedia page with the answer. \\
    & QuAC (Question Answering in Context) & The QuAC benchmark for question answering in the context of dialogues \cite{choi-etal-2018-quac}. \\ 
    \hline
    \multirow{2}{*}{Summarization}
    & XSUM & The XSUM benchmark for text summarization of BBC news articles \cite{narayan-etal-2018-dont}\\
    & CNN/DailyMail & The CNN/DailyMail benchmark for text summarization \cite{nallapati-etal-2016-abstractive}. \\ 
    \hline 
    \end{tabular}
    \end{center}
    \caption{The HELM scenarios we used for our experiments, which include short and long text output tasks.}  \label{table_helm_scenarios}
\end{table*}

%% file: plots/app_fig_all.tex
\begin{figure*}
    \subfloat{\includegraphics[width=0.32\textwidth]{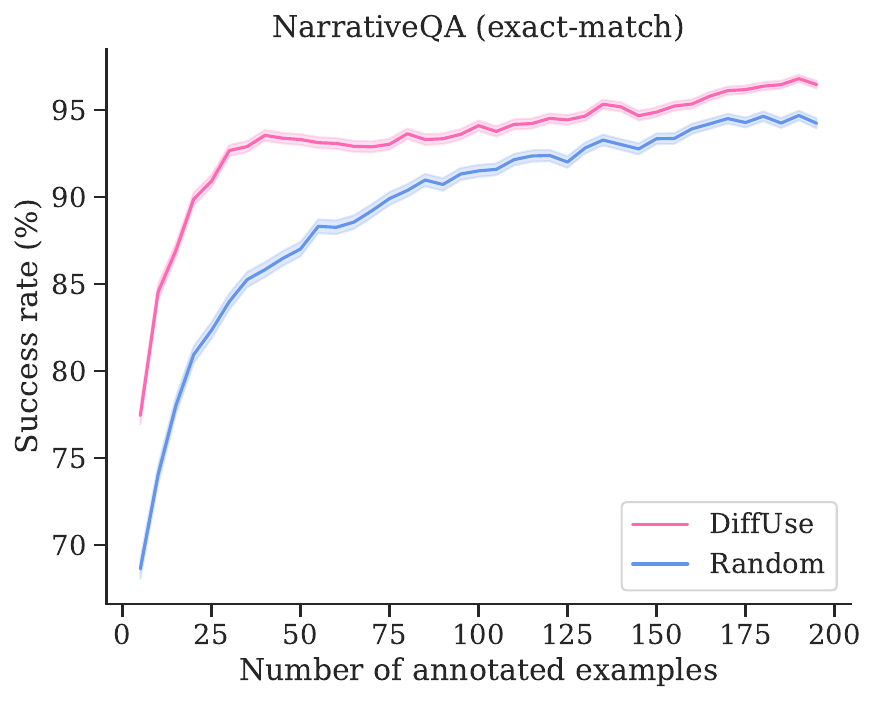}}
    \subfloat{\includegraphics[width=0.32\textwidth]{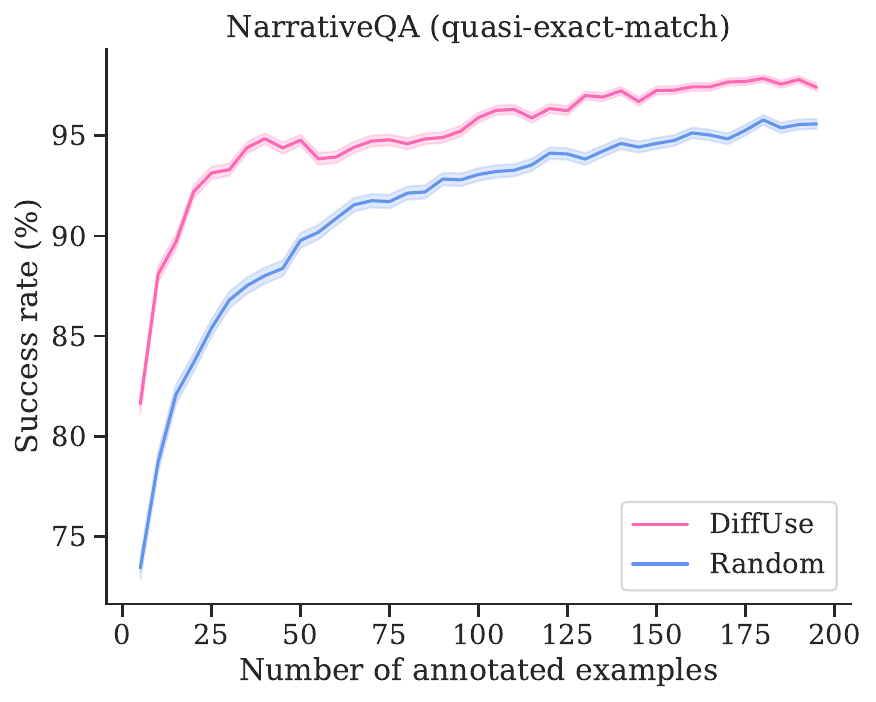}}
    \subfloat{\includegraphics[width=0.32\textwidth]{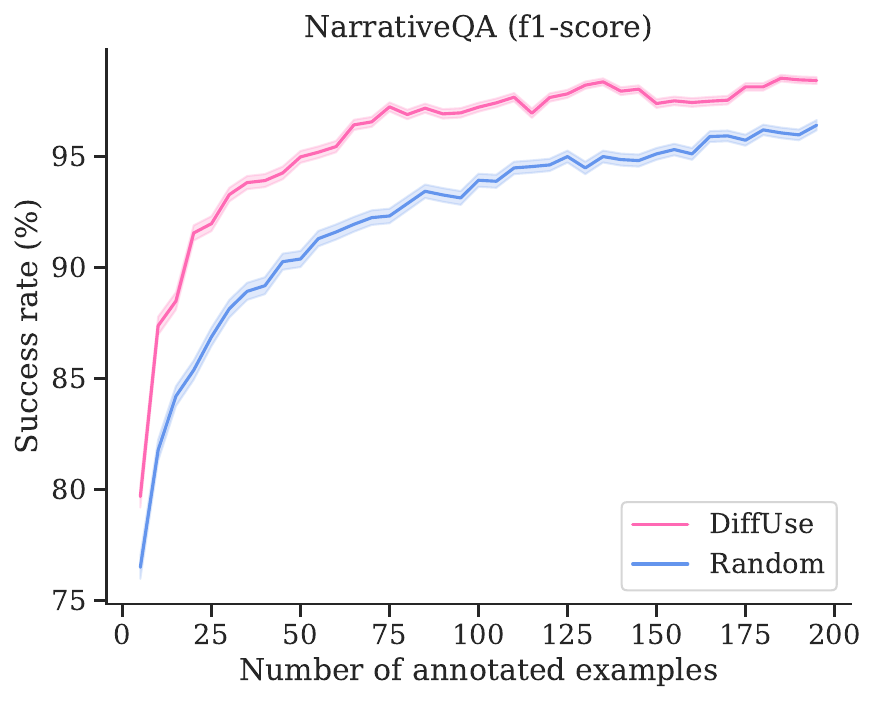}}\\[-2ex]
    \subfloat{\includegraphics[width=0.32\textwidth]{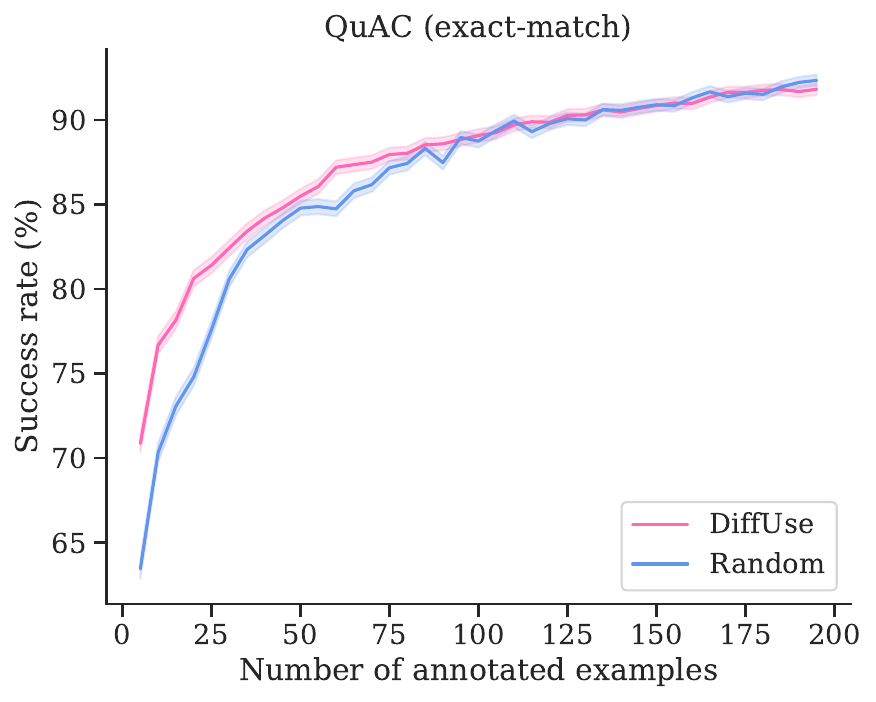}}
    \subfloat{\includegraphics[width=0.32\textwidth]{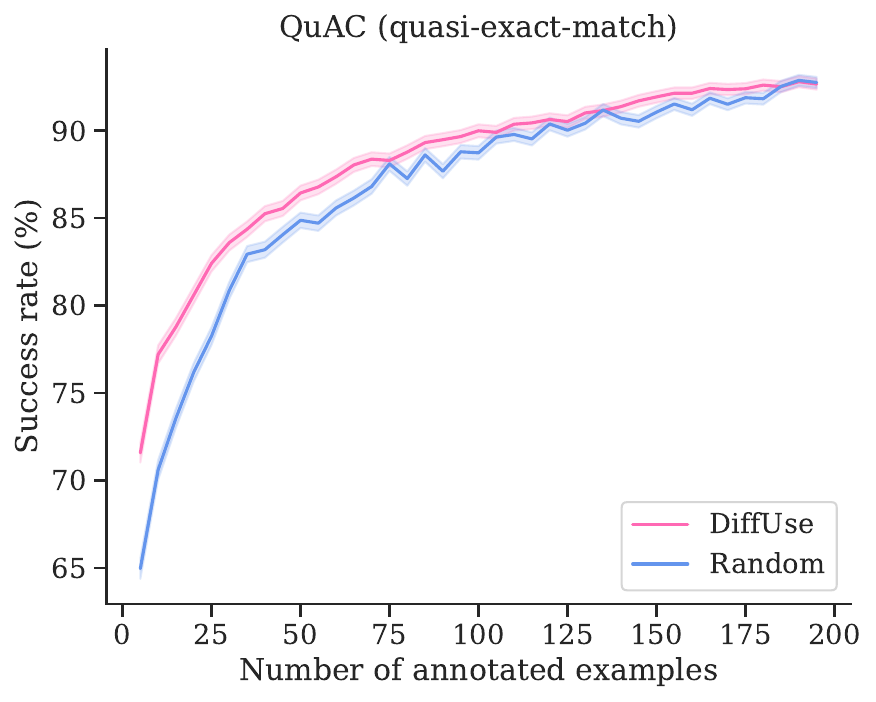}}
    \subfloat{\includegraphics[width=0.32\textwidth]{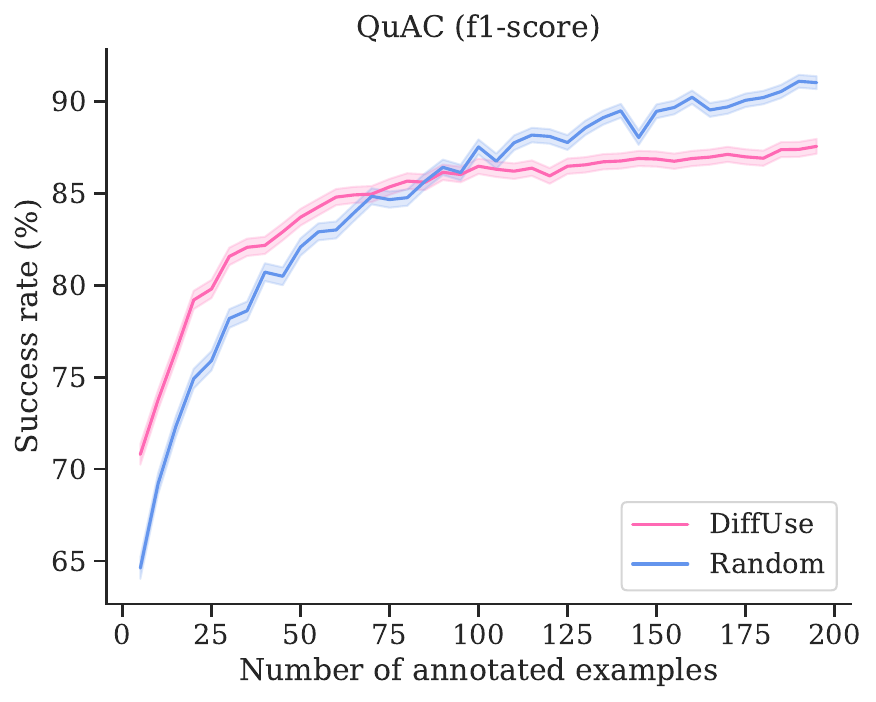}}\\[-2ex]
    \subfloat{\includegraphics[width=0.32\textwidth]{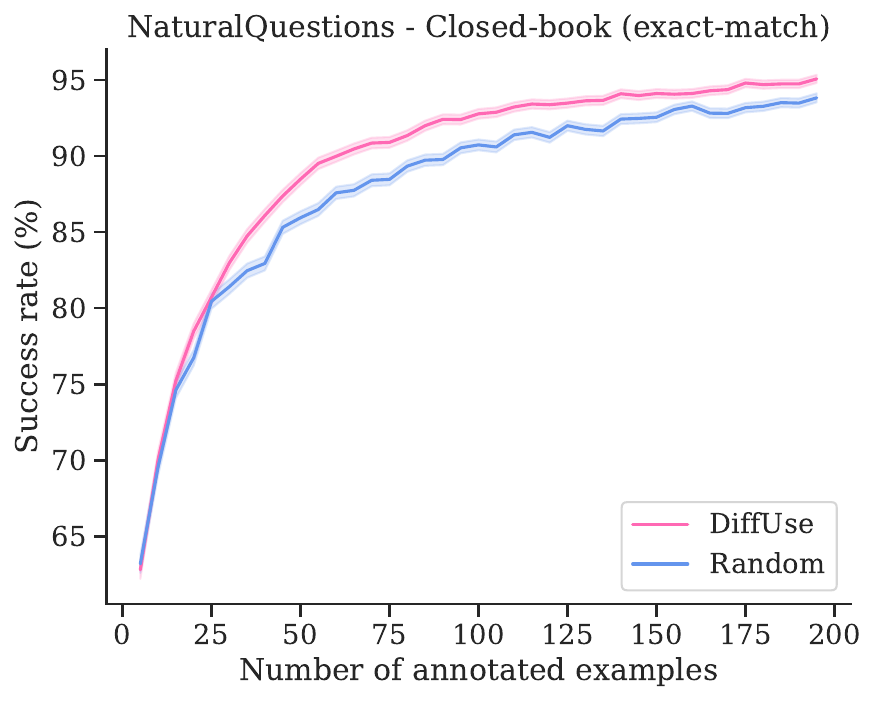}}
    \subfloat{\includegraphics[width=0.32\textwidth]{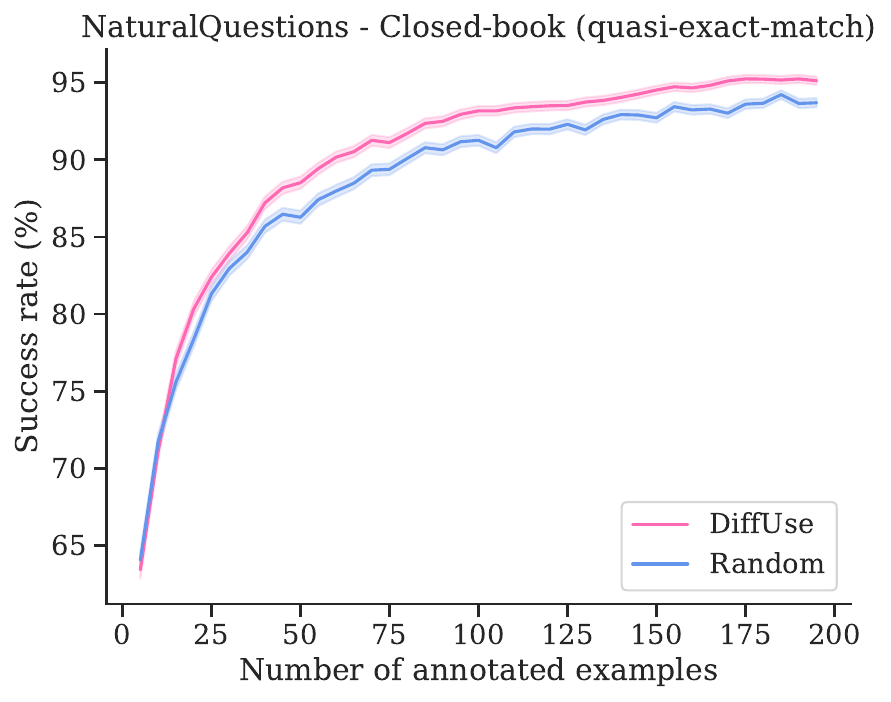}}
    \subfloat{\includegraphics[width=0.32\textwidth]{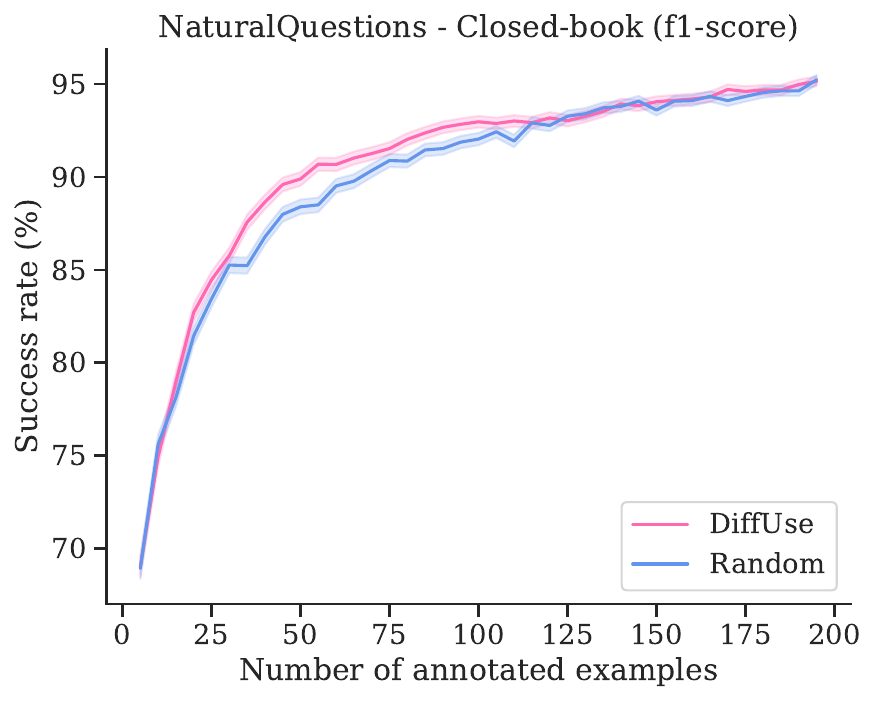}}\\[-2ex]
    \subfloat{\includegraphics[width=0.32\textwidth]{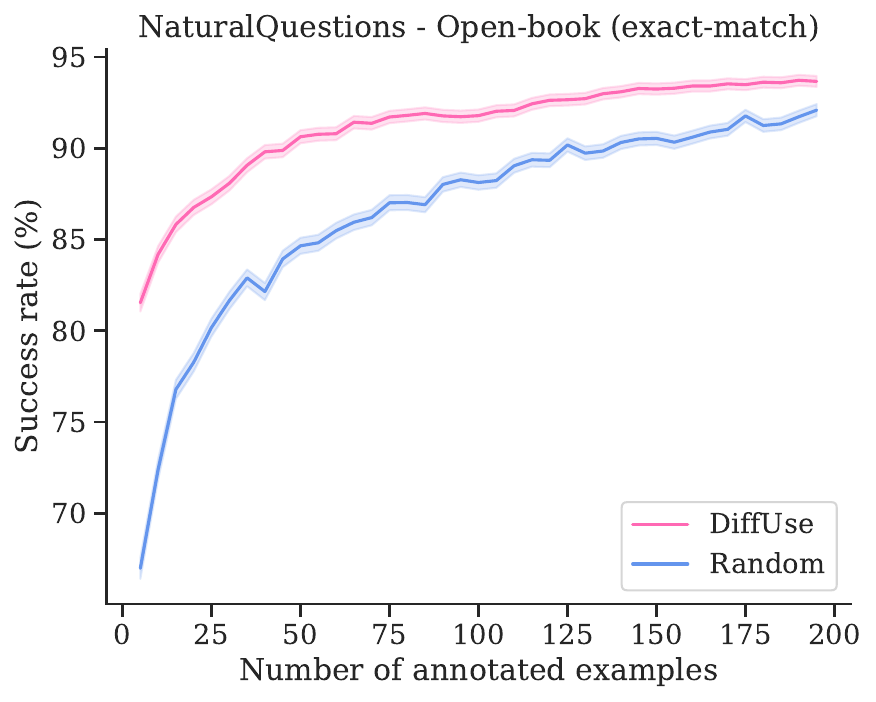}}
    \subfloat{\includegraphics[width=0.32\textwidth]{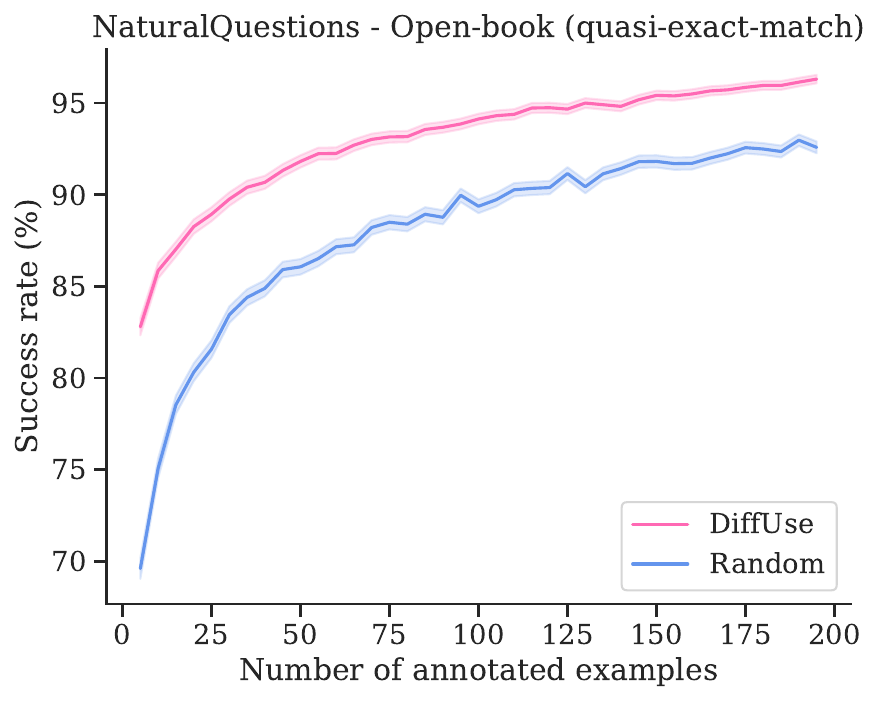}}
    \subfloat{\includegraphics[width=0.32\textwidth]{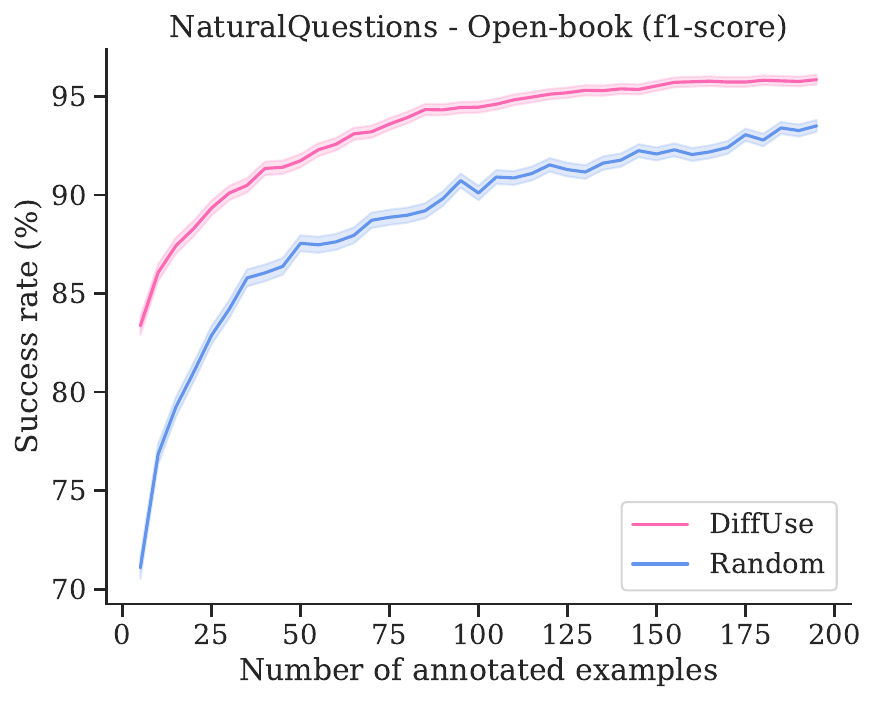}}\\[-2ex]
    \subfloat{\includegraphics[width=0.32\textwidth]{final_plots/main_result/summarization_cnndm_temperature=0.3,device=cuda_rouge_2.pdf}}
    \subfloat{\includegraphics[width=0.32\textwidth]{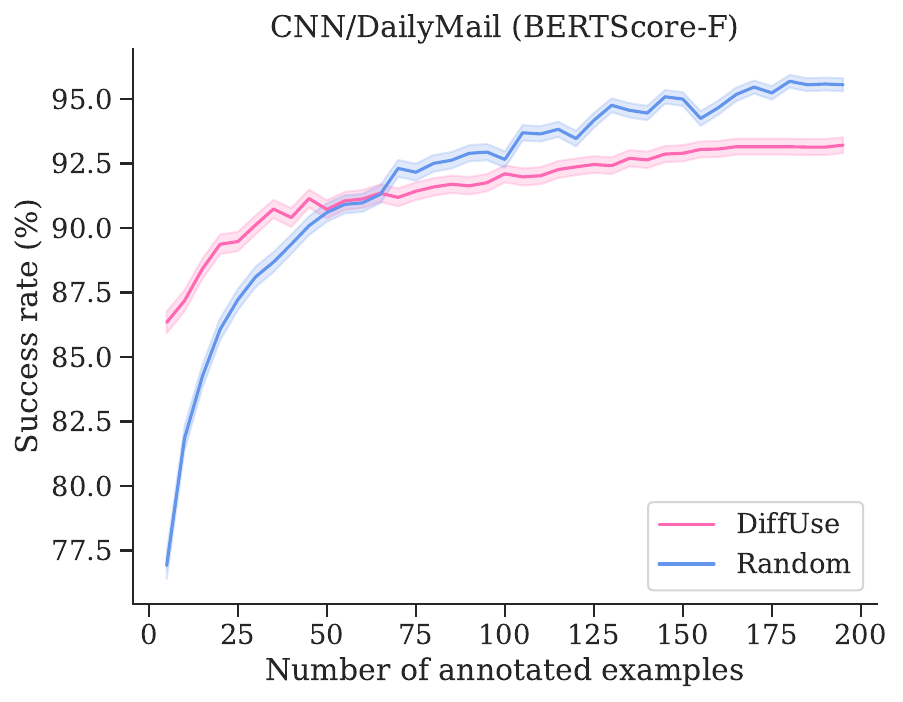}}
    \subfloat{\includegraphics[width=0.32\textwidth]{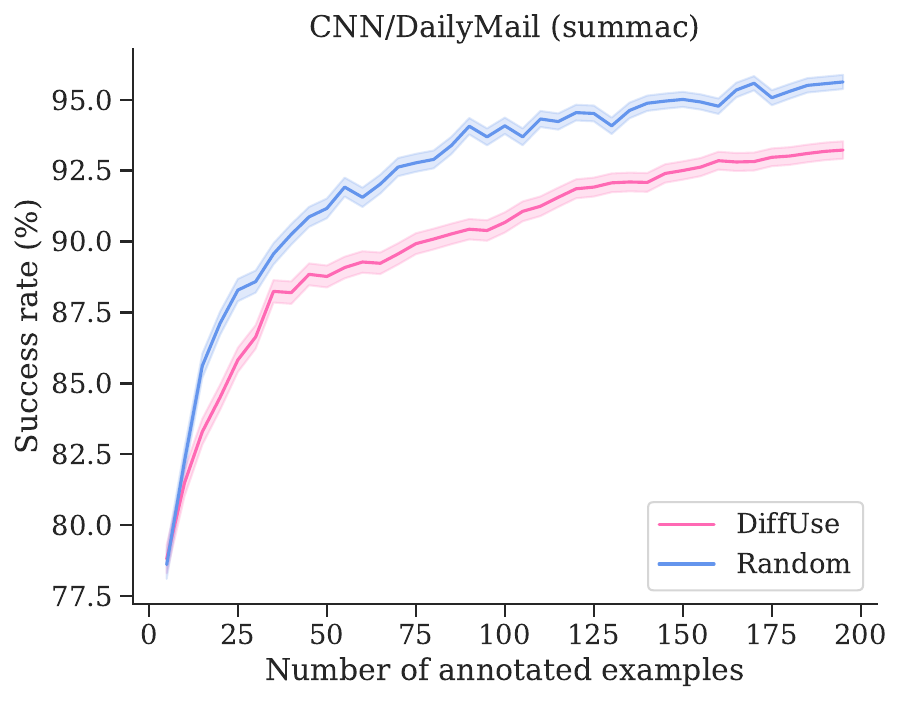}}\\[-2ex]
    \subfloat{\includegraphics[width=0.32\textwidth]{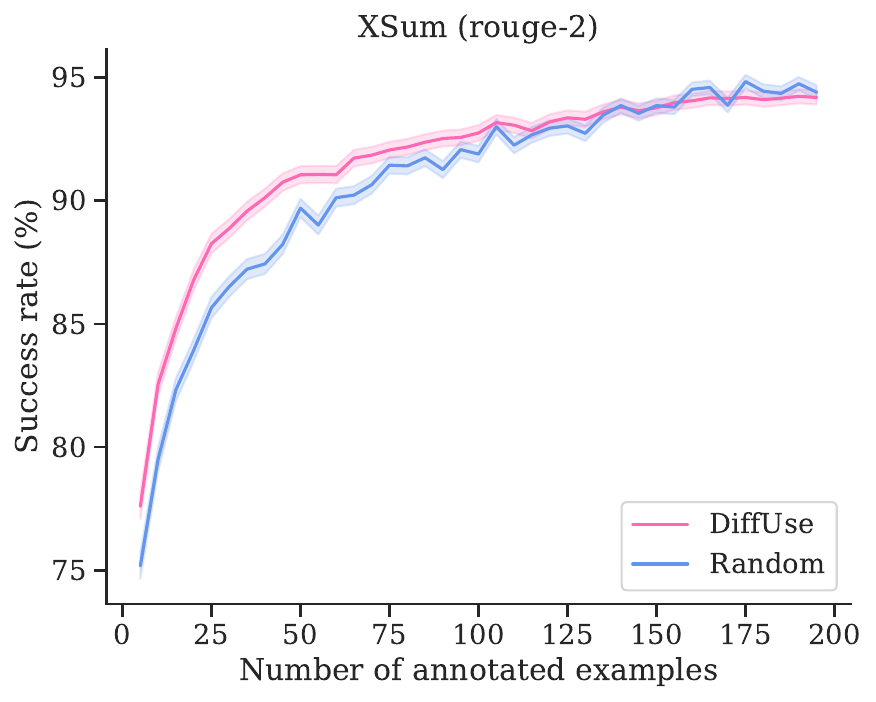}}
    \subfloat{\includegraphics[width=0.32\textwidth]{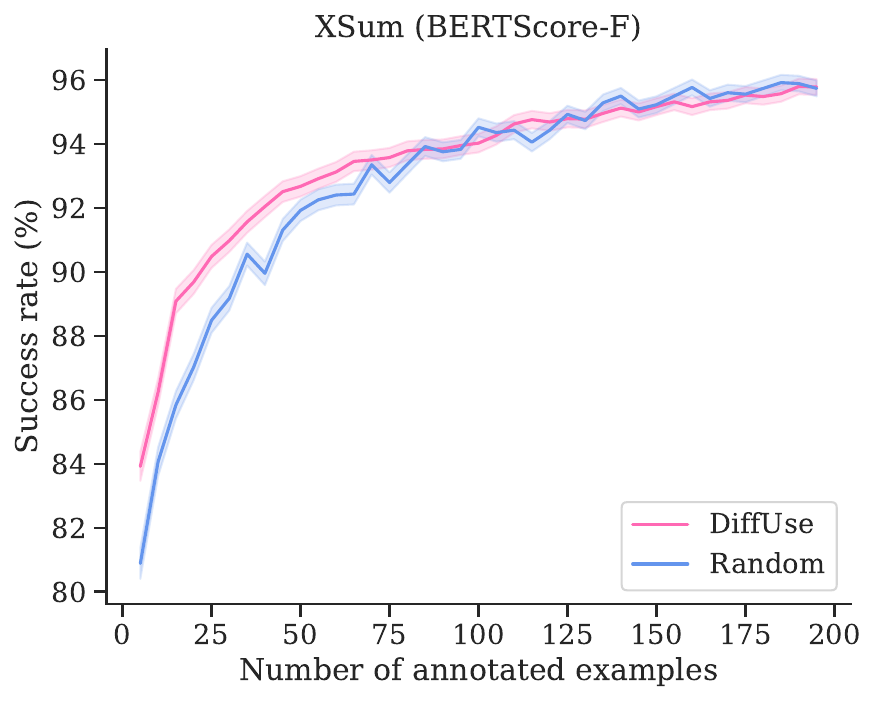}}
    \subfloat{\includegraphics[width=0.32\textwidth]{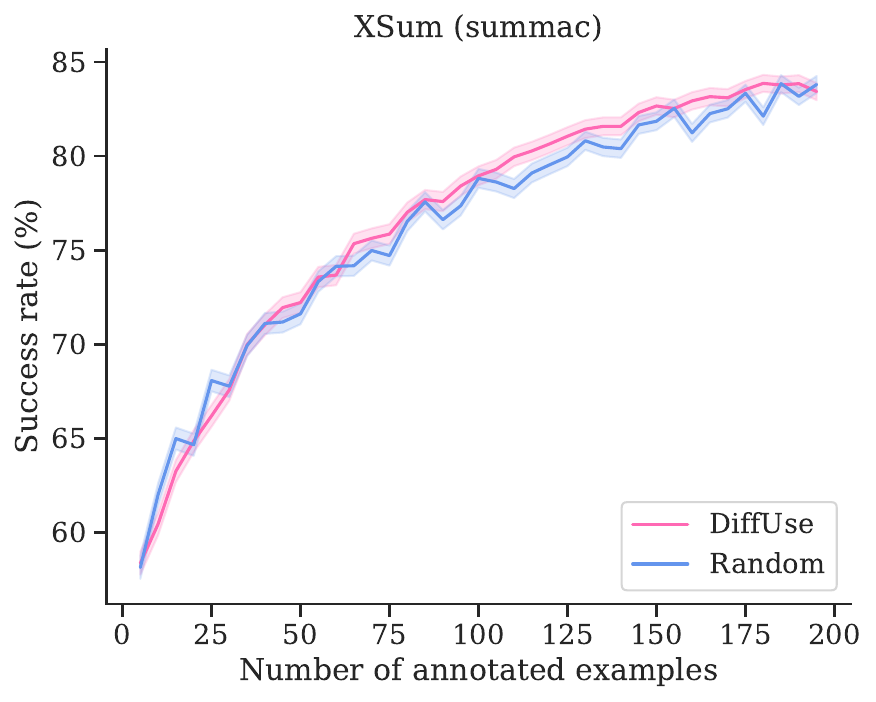}}\\[-2ex]
\caption{\textbf{Full results}. Plots depict success rates of model preference estimation, aggregated over $666$ unique model pairs. Each panel depicts a different combination of dataset and "oracle" (reference-based evaluation metric).}
\label{fig_full_results}
\end{figure*}

%% file: iterative_table.tex
\begin{table*}
\resizebox{\textwidth}{!}{%

\begin{tabular}{llrrrrrrrr}
\toprule
 &  & \# Annotations $\downarrow$ & Error (\%) $\downarrow$ & Success (\%) $\uparrow$ & Inconclusive (\%) $\downarrow$ & Average Distance & Avg. Dist. & Avg. Dist. & Avg. Dist. \\
 Dataset & Method &  &  &  &  &  &   (Error) $\downarrow$ & (Success) & (Inconcl.) $\downarrow$ \\
\midrule
\multirow[t]{2}{*}{CNN/DailyMail} & DiffUse & 15.90 & 11.97 & 86.50 & 1.53 & 0.27 & 0.06 & 0.30 & 0.09 \\
 & Random & 22.47 & 16.77 & 80.32 & 2.91 & 0.27 & 0.10 & 0.31 & 0.07 \\
\cline{1-10}
\multirow[t]{2}{*}{NarrativeQA} & DiffUse & 28.77 & 4.20 & 89.64 & 6.16 & 0.31 & 0.06 & 0.34 & 0.05 \\
 & Random & 118.69 & 0.72 & 43.03 & 56.25 & 0.31 & 0.11 & 0.55 & 0.13 \\
\cline{1-10}
\multirow[t]{2}{*}{NaturalQuestions
Closed-book} & DiffUse & 189.05 & 0.20 & 6.40 & 93.41 & 0.15 & 0.12 & 0.32 & 0.13 \\
 & Random & 191.74 & 0.11 & 4.37 & 95.53 & 0.15 & 0.10 & 0.32 & 0.14 \\
\cline{1-10}
\multirow[t]{2}{*}{NaturalQuestions
Open-book} & DiffUse & 85.25 & 0.93 & 62.81 & 36.26 & 0.20 & 0.03 & 0.28 & 0.06 \\
 & Random & 160.97 & 0.17 & 21.79 & 78.05 & 0.20 & 0.15 & 0.44 & 0.13 \\
\cline{1-10}
\multirow[t]{2}{*}{QuAC} & DiffUse & 86.91 & 7.75 & 58.24 & 34.01 & 0.15 & 0.07 & 0.20 & 0.08 \\
 & Random & 128.93 & 4.37 & 35.93 & 59.70 & 0.15 & 0.09 & 0.25 & 0.10 \\
\cline{1-10}
\multirow[t]{2}{*}{XSum} & DiffUse & 44.25 & 6.86 & 79.38 & 13.75 & 0.30 & 0.09 & 0.35 & 0.08 \\
 & Random & 59.39 & 6.05 & 72.45 & 21.50 & 0.30 & 0.10 & 0.37 & 0.09 \\
\cline{1-10}
\bottomrule
\end{tabular}
}
\caption{\textbf{Iterative selection results} ($p=0.2$). The table depicts the results of applying iterative selection (Algorithm~\ref{algo:model_comparison}; with $p=0.2$, $n=5$, and $N=200$), comparing \method{} to random sampling. Results are aggregated across $666$ model pairs. The table details the amount of annotations performed before reaching the stopping criterion, and the outcomes of the selection experiments (\textit{Success/Error/Inconclusive}). In addition, it details the average winning distance (\S\ref{sec:problem_formulation}) between model pairs, broken down by the experiment outcomes. $\downarrow$: Lower is better. \\
Where the experiment result is inconclusive or the wrong winning model is chosen, the performance gap between models is quite small; Thus, even where the user is unable to correctly determine the better-performing model, the cost of this failure is relatively limited.}
\label{table_iterative}
\end{table*}

\begin{table*}
\resizebox{\textwidth}{!}{%

\begin{tabular}{llrrrrrrrr}
\toprule
 &  & \# Annotations $\downarrow$ & Error (\%) $\downarrow$ & Success (\%) $\uparrow$ & Inconclusive (\%) $\downarrow$ & Average Distance & Avg. Dist. & Avg. Dist. & Avg. Dist. \\
 Dataset & Method &  &  &  &  &  &   (Error) $\downarrow$ & (Success) & (Inconcl.) $\downarrow$ \\
\midrule
\multirow[t]{2}{*}{CNN/DailyMail} & DiffUse & 34.54 & 7.13 & 86.44 & 6.43 & 0.27 & 0.05 & 0.30 & 0.07 \\
 & Random & 51.98 & 8.65 & 79.32 & 12.03 & 0.27 & 0.07 & 0.32 & 0.07 \\
\cline{1-10}
\multirow[t]{2}{*}{NarrativeQA} & DiffUse & 47.80 & 1.35 & 84.53 & 14.11 & 0.31 & 0.05 & 0.36 & 0.05 \\
 & Random & 132.75 & 0.05 & 37.69 & 62.27 & 0.31 & 0.05 & 0.59 & 0.14 \\
\cline{1-10}
\multirow[t]{2}{*}{NaturalQuestions
Closed-book} & DiffUse & 195.37 & 0.00 & 3.30 & 96.70 & 0.15 & NaN & 0.37 & 0.14 \\
 & Random & 197.35 & 0.00 & 1.65 & 98.35 & 0.15 & NaN & 0.40 & 0.14 \\
\cline{1-10}
\multirow[t]{2}{*}{NaturalQuestions
Open-book} & DiffUse & 111.02 & 0.23 & 52.87 & 46.91 & 0.20 & 0.02 & 0.30 & 0.07 \\
 & Random & 172.34 & 0.02 & 16.71 & 83.27 & 0.20 & 0.05 & 0.49 & 0.14 \\
\cline{1-10}
\multirow[t]{2}{*}{QuAC} & DiffUse & 129.20 & 2.96 & 43.26 & 53.78 & 0.15 & 0.06 & 0.23 & 0.09 \\
 & Random & 158.26 & 1.14 & 25.11 & 73.75 & 0.15 & 0.08 & 0.29 & 0.11 \\
\cline{1-10}
\multirow[t]{2}{*}{XSum} & DiffUse & 71.18 & 2.16 & 73.21 & 24.62 & 0.30 & 0.07 & 0.37 & 0.08 \\
 & Random & 88.84 & 2.09 & 63.90 & 34.01 & 0.30 & 0.08 & 0.41 & 0.10 \\
\cline{1-10}
\bottomrule
\end{tabular}
}
\caption{\textbf{Iterative selection results} ($p=0.1$). The table depicts the results of applying iterative selection (Algorithm~\ref{algo:model_comparison}; with $p=0.1$, $n=5$, and $N=200$), comparing \method{} to random sampling. Results are aggregated across $666$ model pairs. The table details the amount of annotations performed before reaching the stopping criterion, and the outcomes of the selection experiments (\textit{Success/Error/Inconclusive}). In addition, it details the average winning distance (\S\ref{sec:problem_formulation}) between model pairs, broken down by the experiment outcomes. $\downarrow$: Lower is better. \\
Where the experiment result is inconclusive or the wrong winning model is chosen, the performance gap between models is quite small; Thus, even where the user is unable to correctly determine the better-performing model, the cost of this failure is relatively limited.}
\label{table_iterative_0.1}
\end{table*}

%% file: plots/app_fig_clustering.tex
\begin{figure*}
    \subfloat{\includegraphics[width=0.32\textwidth]{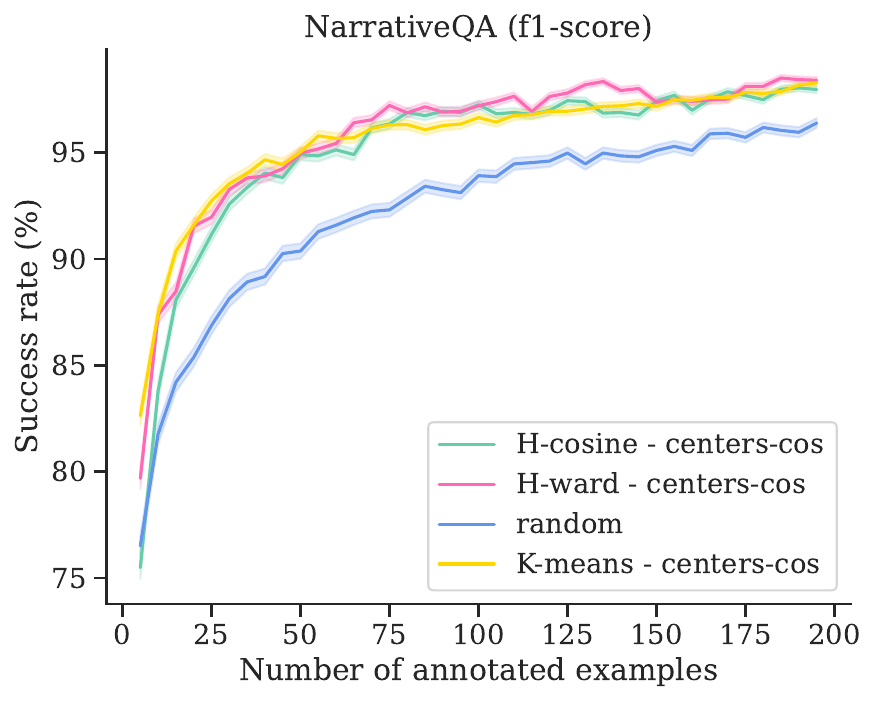}}
    \subfloat{\includegraphics[width=0.32\textwidth]{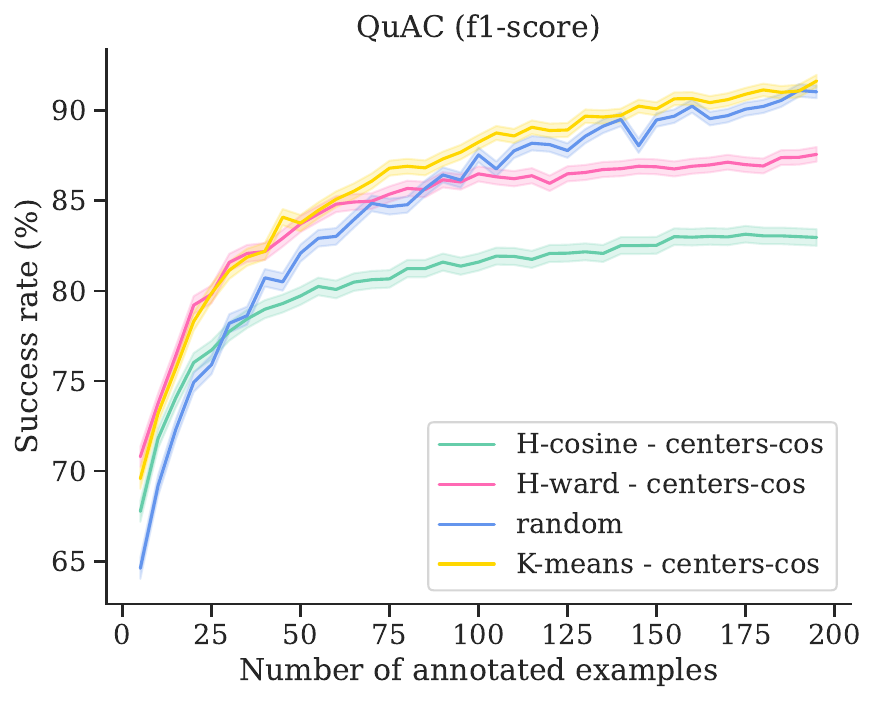}}
    \subfloat{\includegraphics[width=0.32\textwidth]{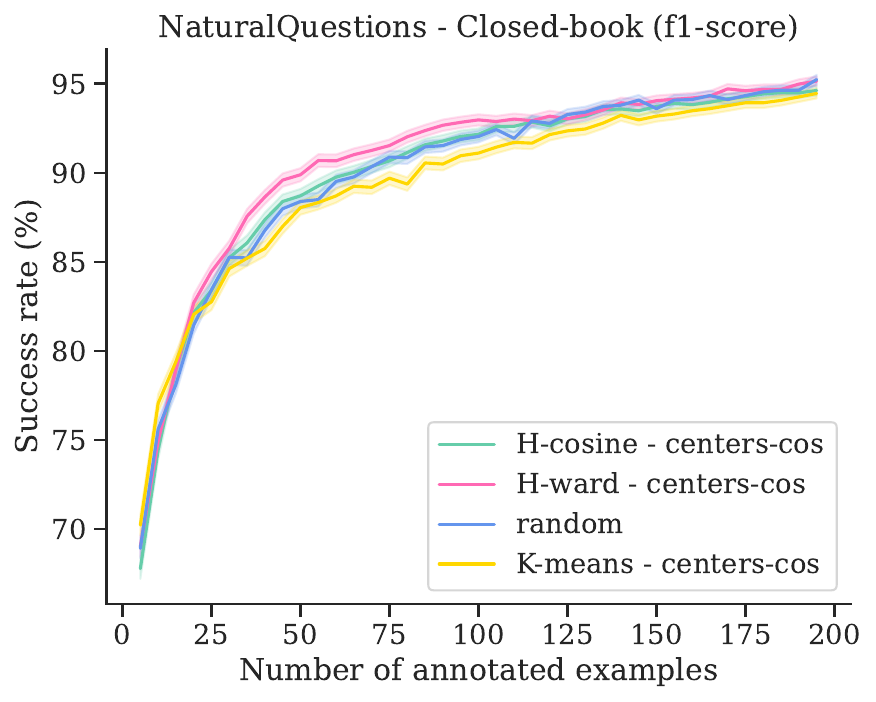}}\\[-2ex]
    \subfloat{\includegraphics[width=0.32\textwidth]{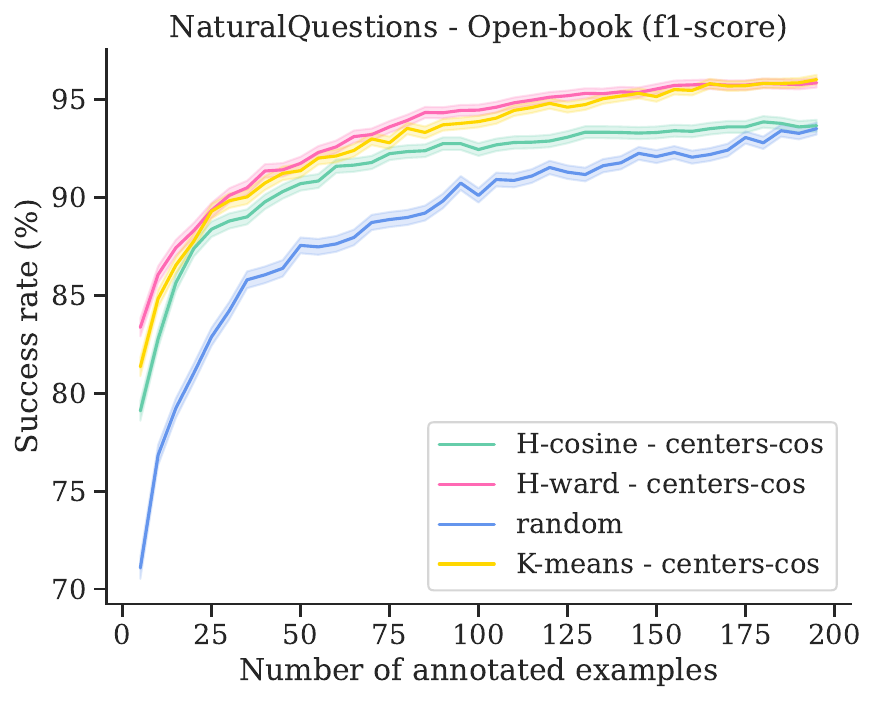}}
    \subfloat{\includegraphics[width=0.32\textwidth]{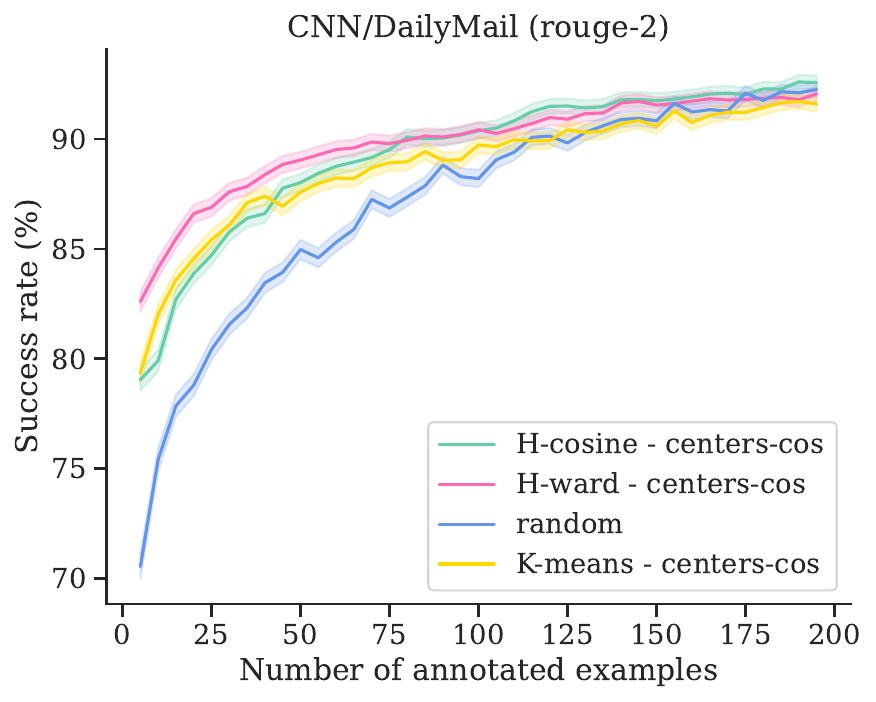}}
    \subfloat{\includegraphics[width=0.32\textwidth]{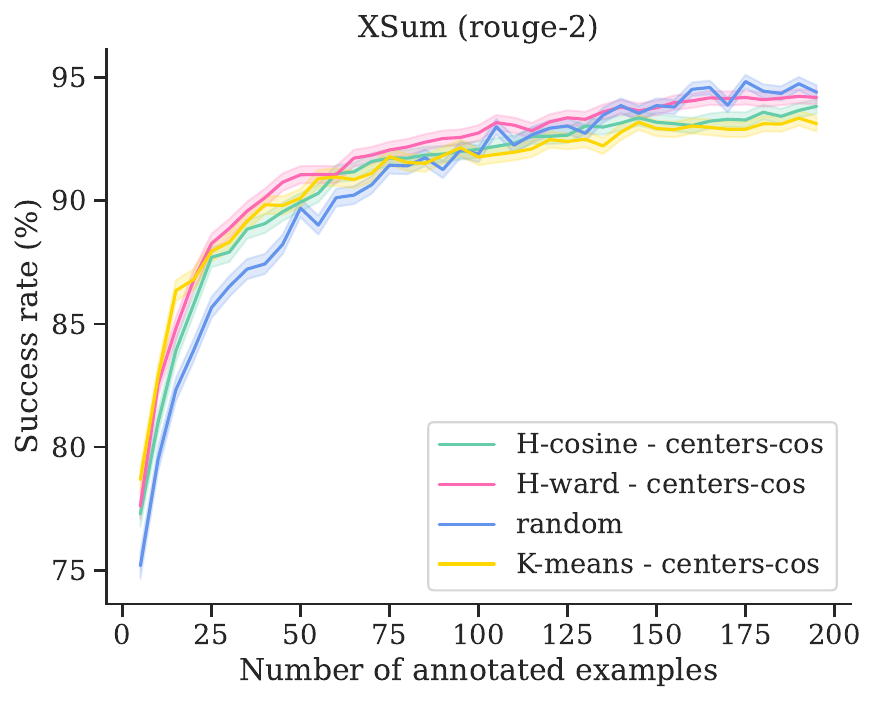}}
\caption{\textbf{Comparing clustering algorithms}. Plots depict success rates of model preference estimation, aggregated over $666$ unique model pairs. Each panel depicts a different dataset. For all clustering methods, a single example -- closest in cosine distance to the cluster center -- is selected from each cluster.}
\label{fig_clustering}
\end{figure*}

%% file: plots/app_fig_representatives.tex
\begin{figure*}
    \subfloat{\includegraphics[width=0.32\textwidth]{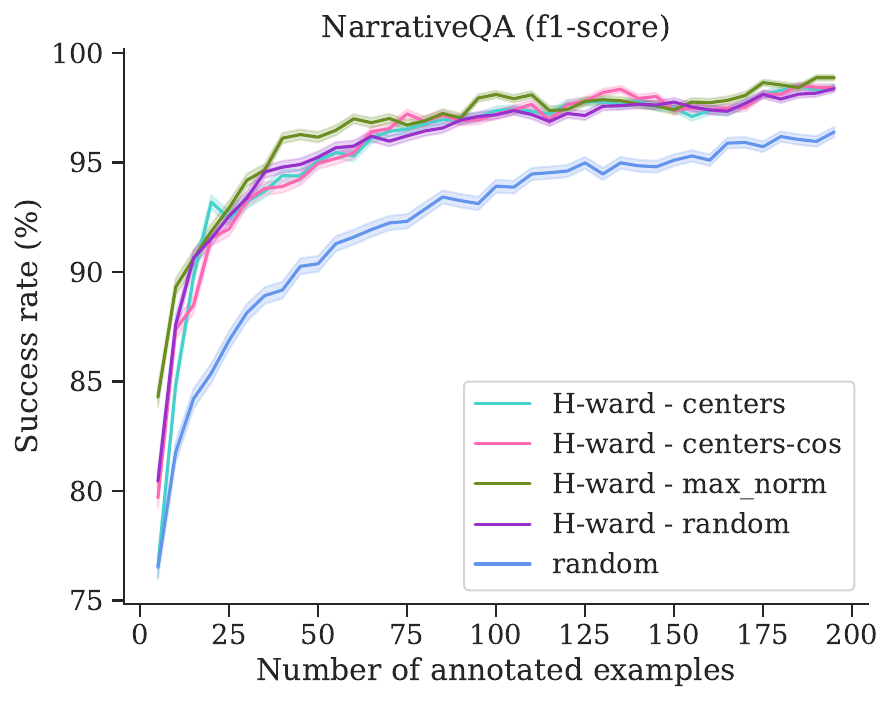}}
    \subfloat{\includegraphics[width=0.32\textwidth]{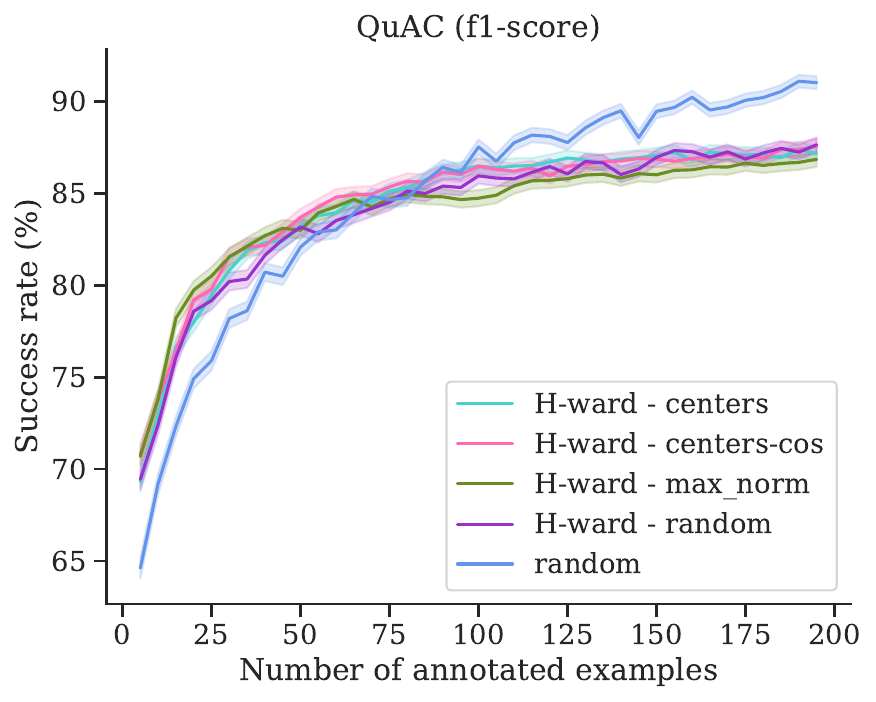}}
    \subfloat{\includegraphics[width=0.32\textwidth]{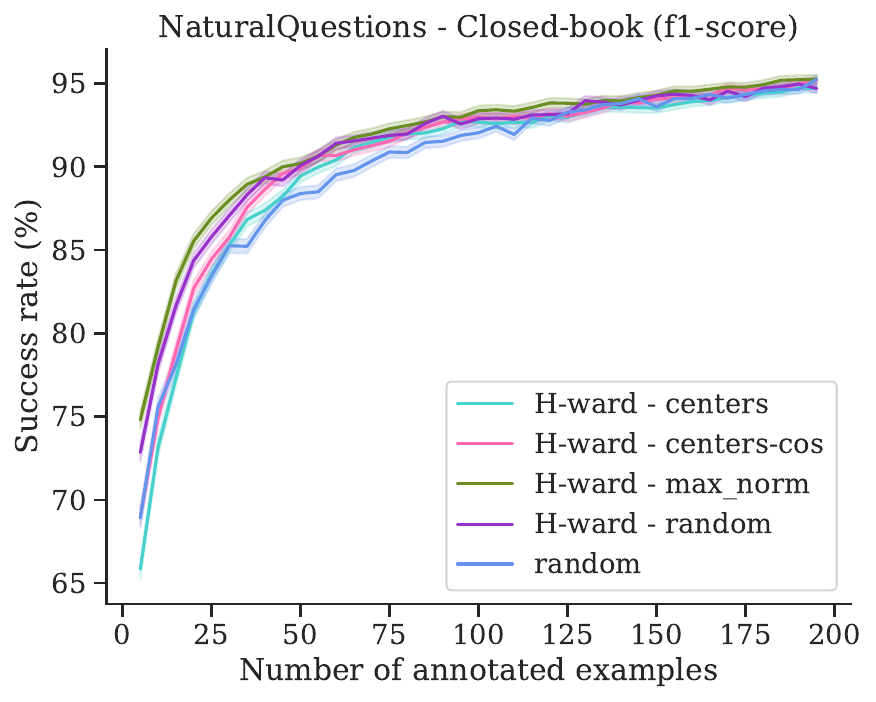}}\\[-2ex]
    \subfloat{\includegraphics[width=0.32\textwidth]{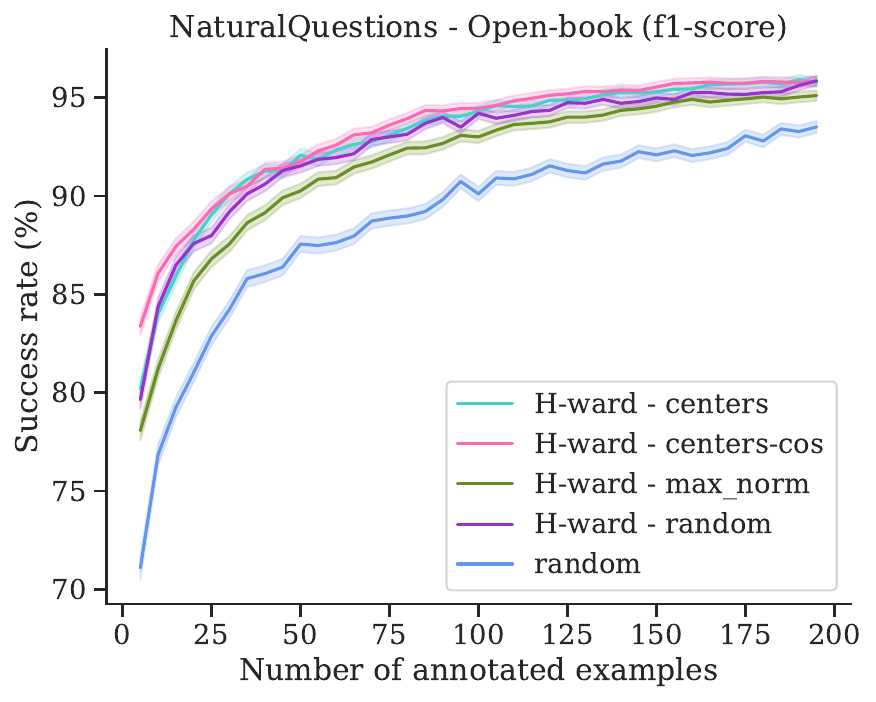}}
    \subfloat{\includegraphics[width=0.32\textwidth]{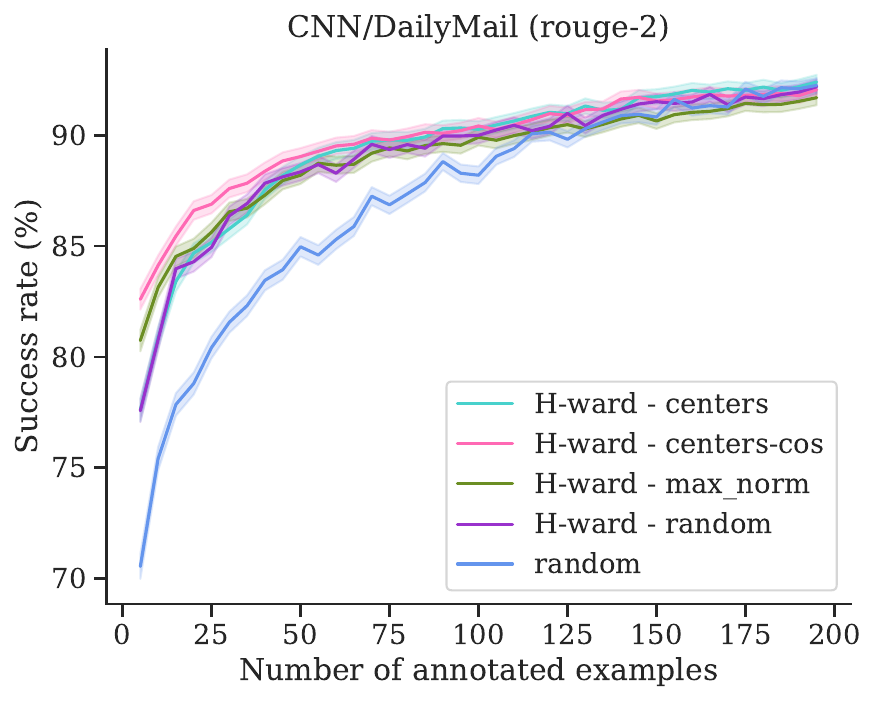}}
    \subfloat{\includegraphics[width=0.32\textwidth]{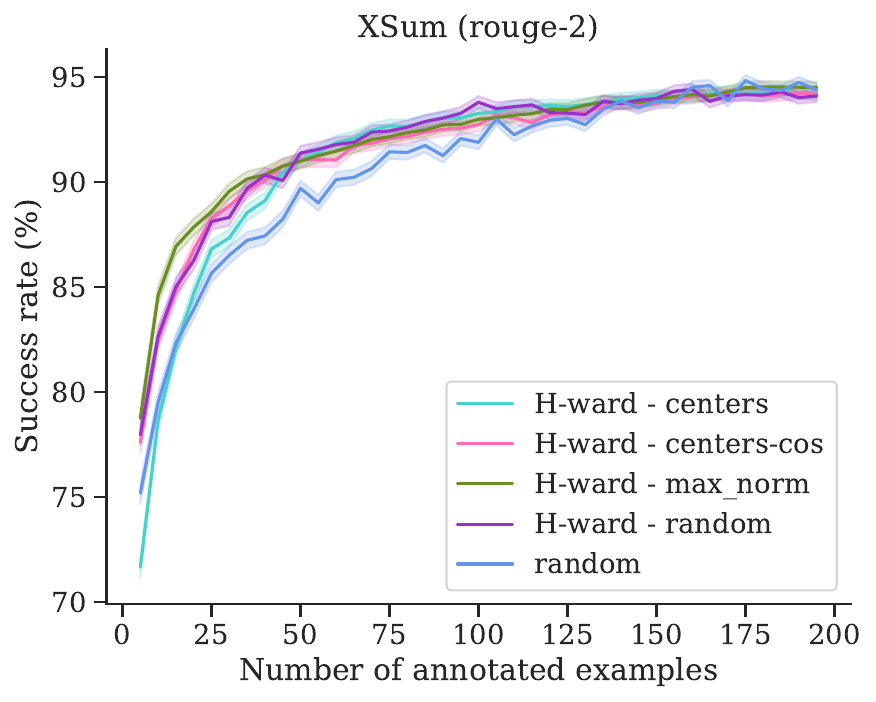}}
\caption{\textbf{Comparing representative selection methods}. Plots depict success rates of model preference estimation, aggregated over $666$ unique model pairs. Each panel depicts a different dataset. For all non-random methods, hierarchical clustering with Ward linkage was used to partition the difference vectors; the plots compare approaches for selecting a single representative from each cluster.}
\label{fig_representatives}
\end{figure*}

%% file: plots/app_fig_inputs.tex
\begin{figure*}
    \subfloat{\includegraphics[width=0.32\textwidth]{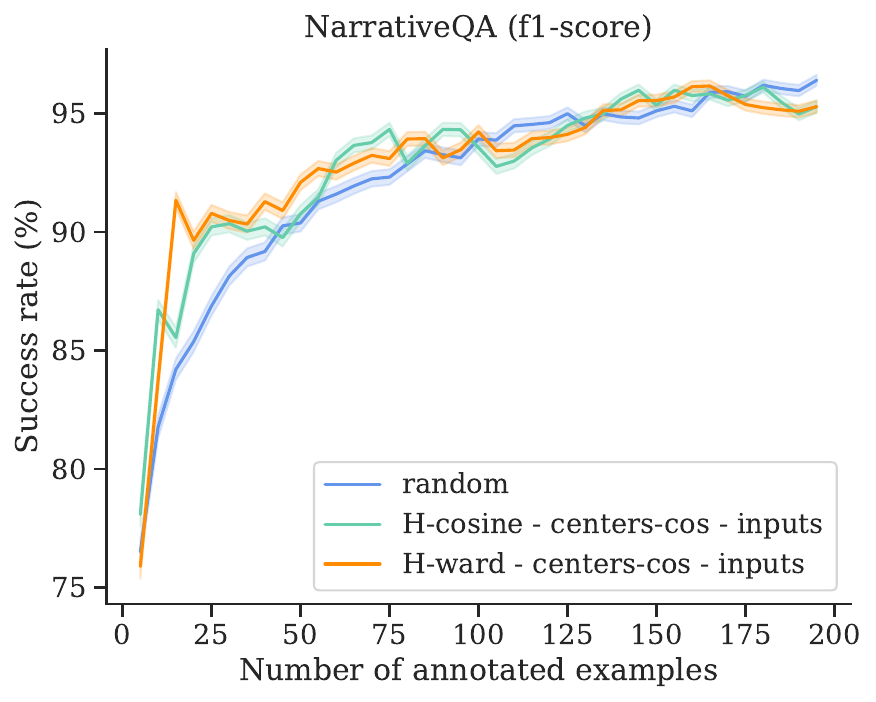}}
    \subfloat{\includegraphics[width=0.32\textwidth]{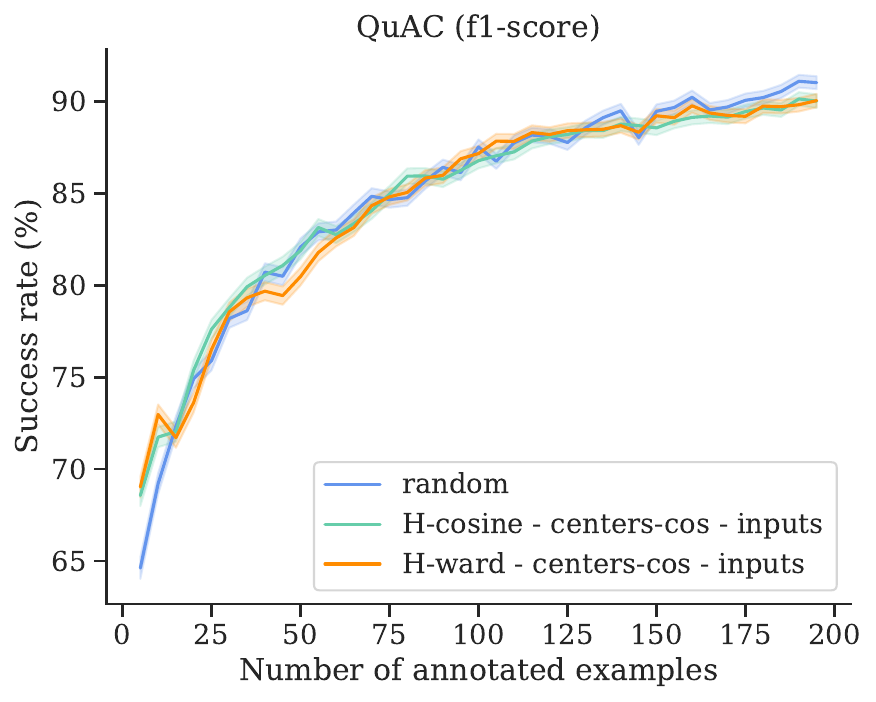}}
    \subfloat{\includegraphics[width=0.32\textwidth]{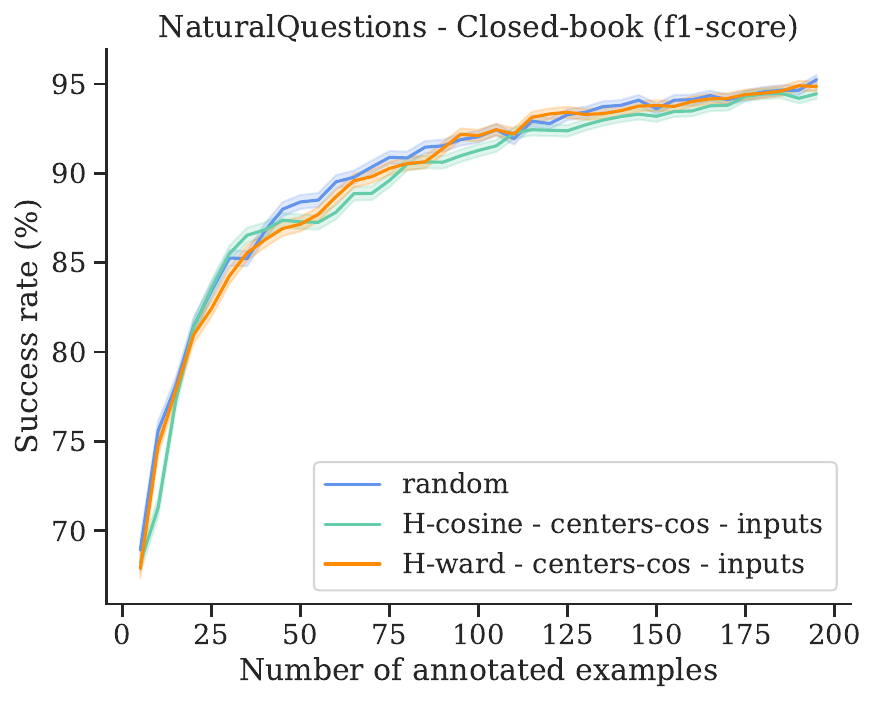}}\\[-2ex]
    \subfloat{\includegraphics[width=0.32\textwidth]{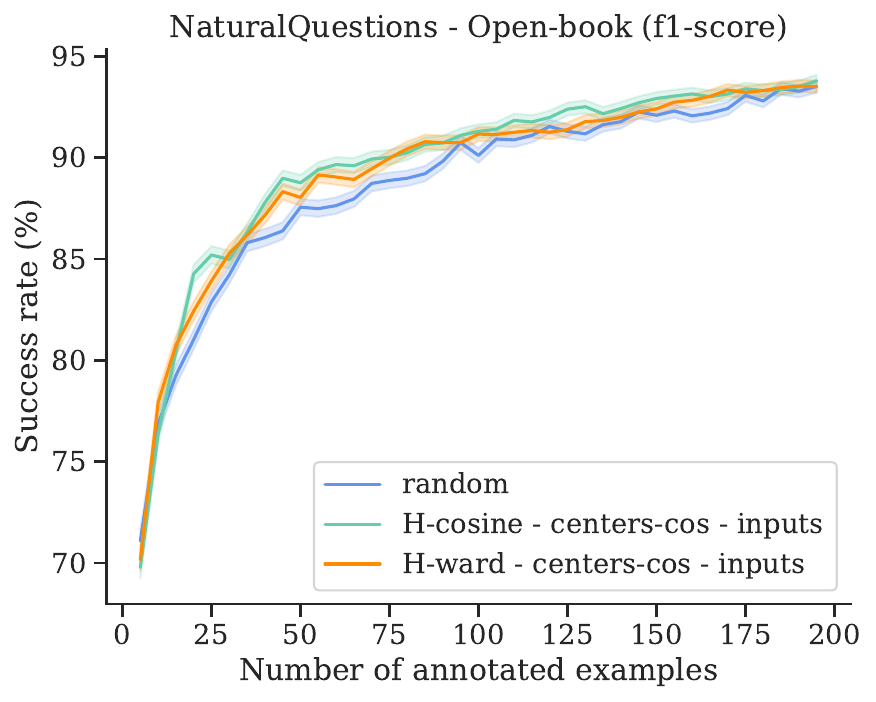}}
    \subfloat{\includegraphics[width=0.32\textwidth]{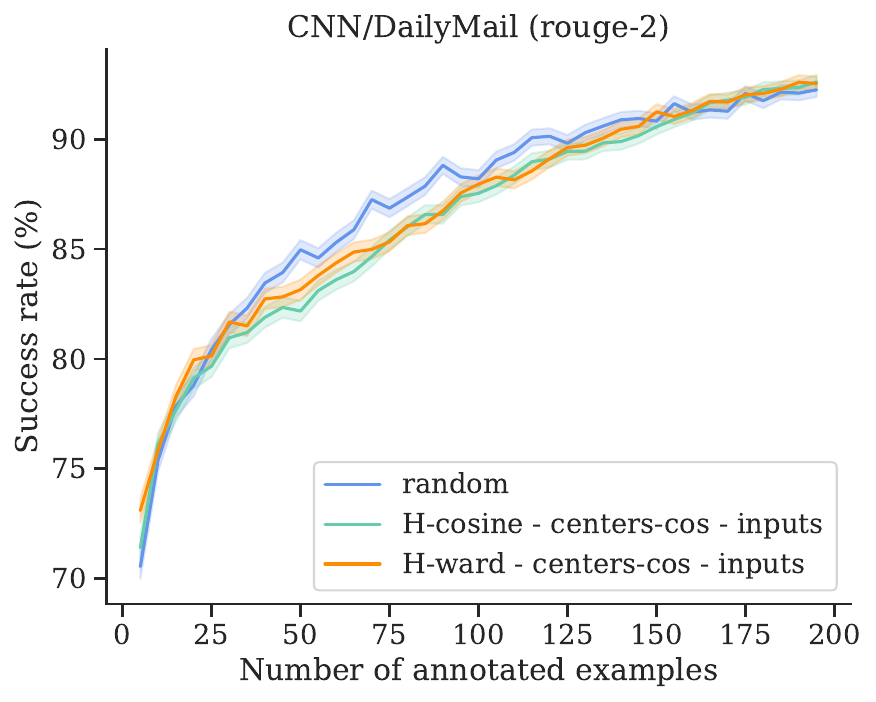}}
    \subfloat{\includegraphics[width=0.32\textwidth]{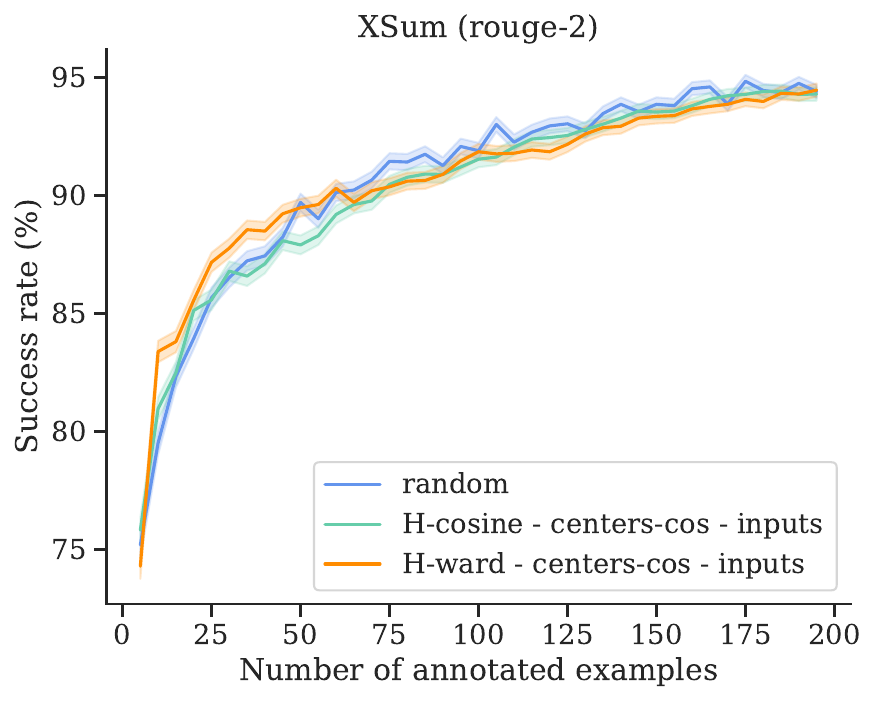}}
\caption{\textbf{Input-based clustering results}. Plots depict success rates of model preference estimation, aggregated over $666$ unique model pairs. Each panel depicts a different dataset.}
\label{fig_inputs_baseline}
\end{figure*}

%% file: plots/app_fig_max_norm.tex
\begin{figure*}
    \subfloat{\includegraphics[width=0.32\textwidth]{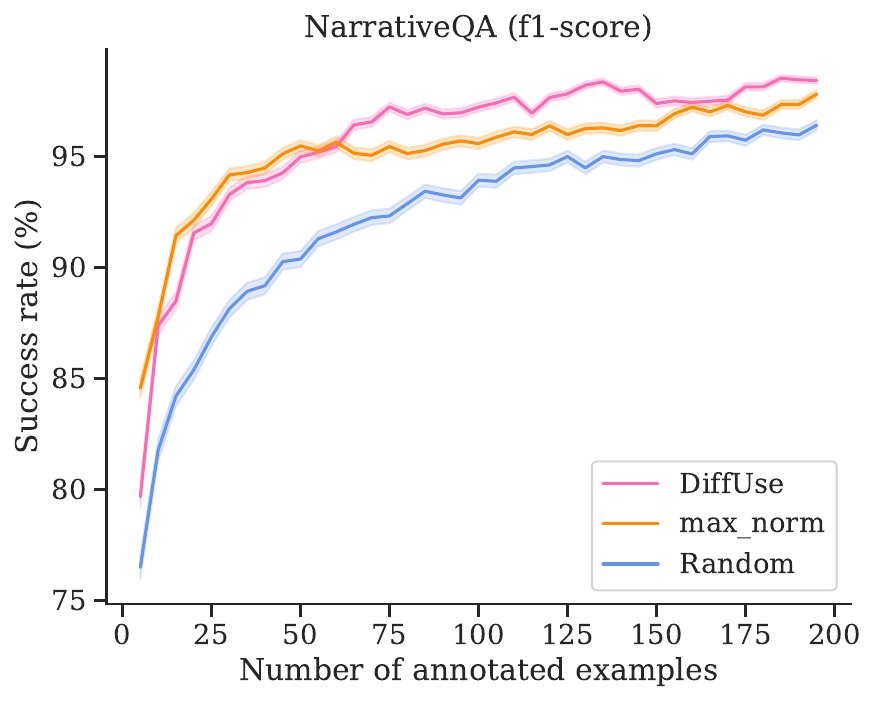}}
    \subfloat{\includegraphics[width=0.32\textwidth]{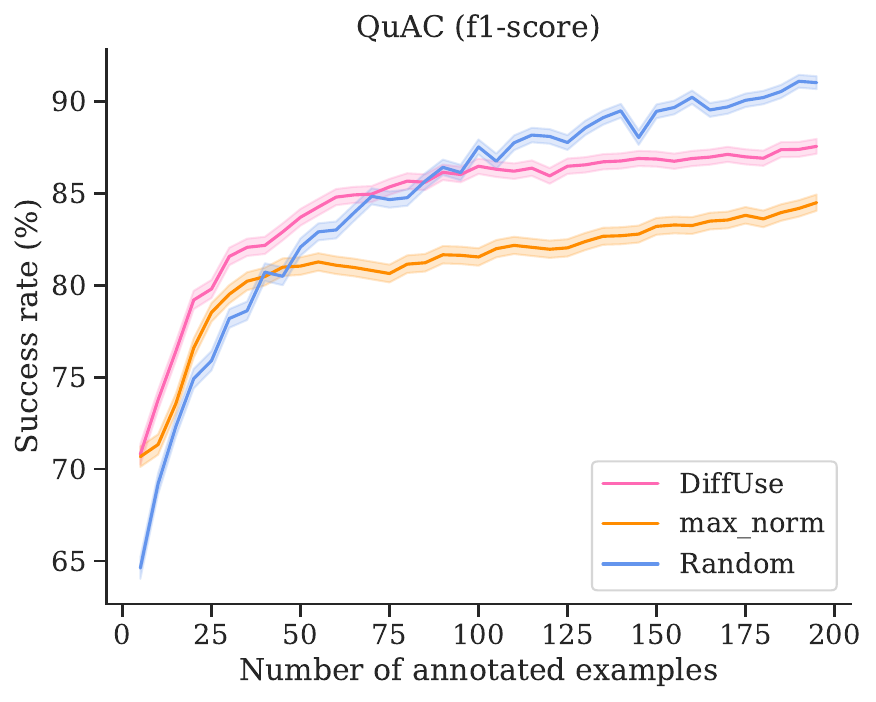}}
    \subfloat{\includegraphics[width=0.32\textwidth]{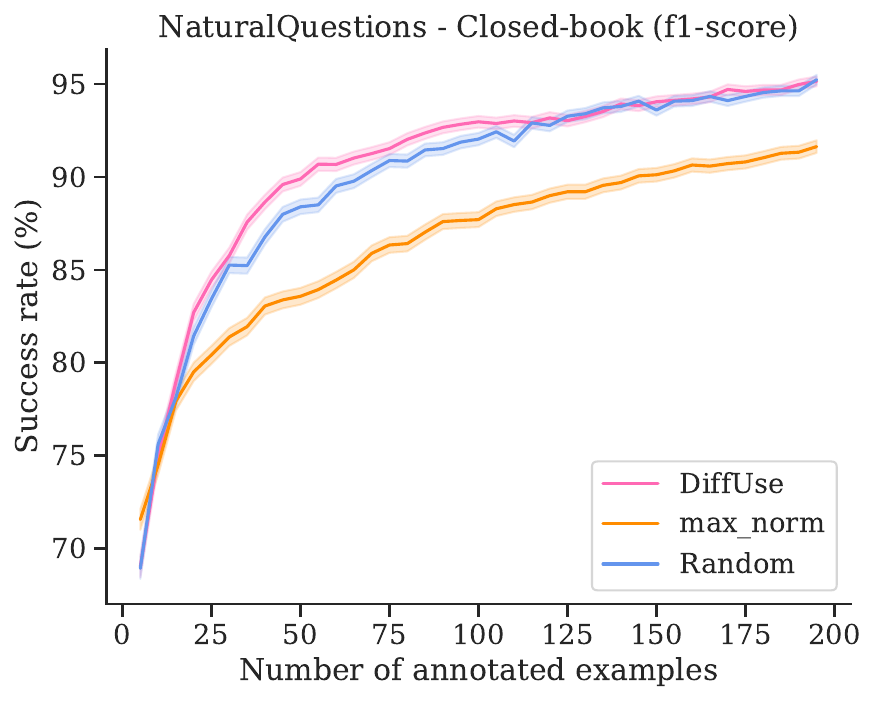}}\\[-2ex]
    \subfloat{\includegraphics[width=0.32\textwidth]{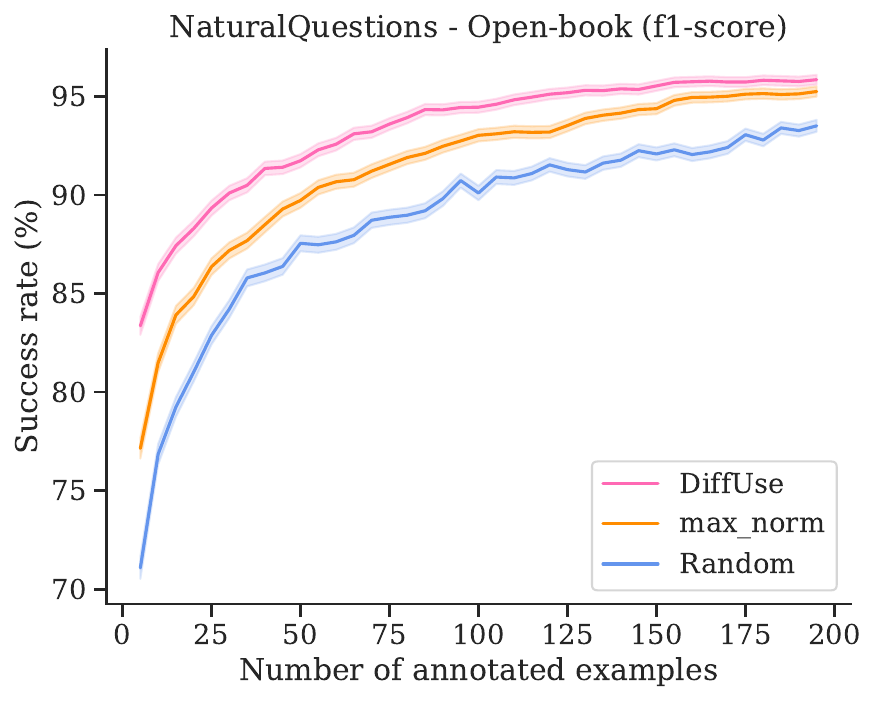}}
    \subfloat{\includegraphics[width=0.32\textwidth]{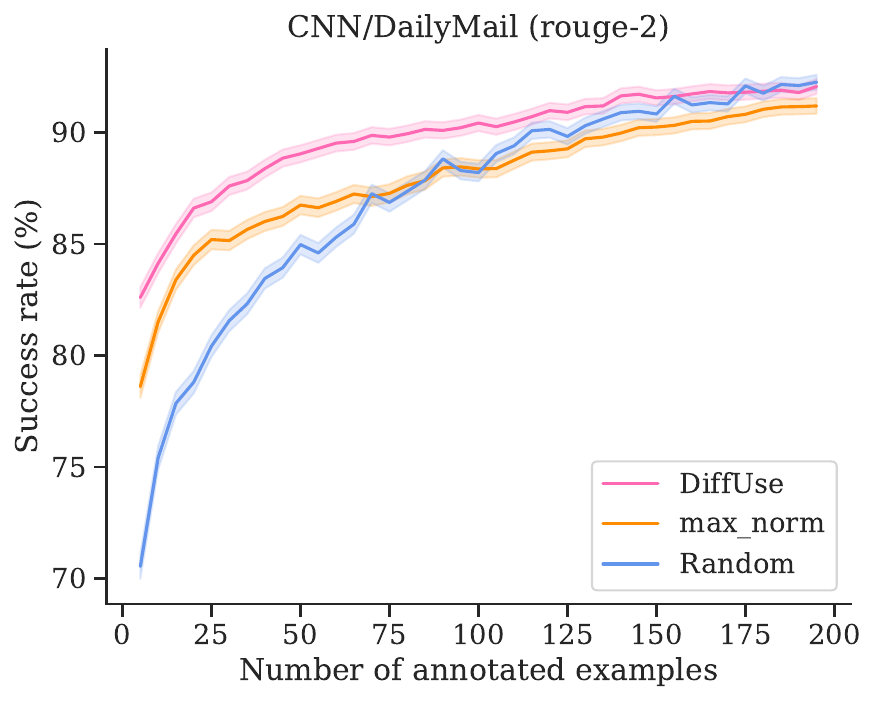}}
    \subfloat{\includegraphics[width=0.32\textwidth]{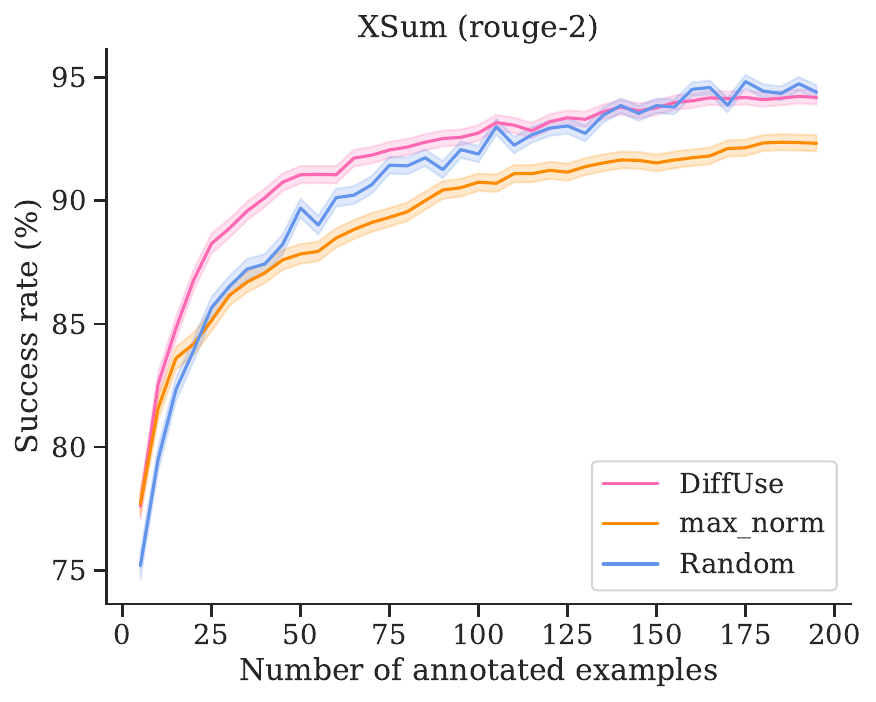}}
\caption{\textbf{Max-norm baseline results}. Plots depict success rates of model preference estimation, aggregated over $666$ unique model pairs. Each panel depicts a different dataset.}
\label{graphs_norm_baseline}
\end{figure*}

%% file: plots/app_fig_prompts.tex
\begin{figure*}
    \subfloat{\includegraphics[width=0.32\textwidth]{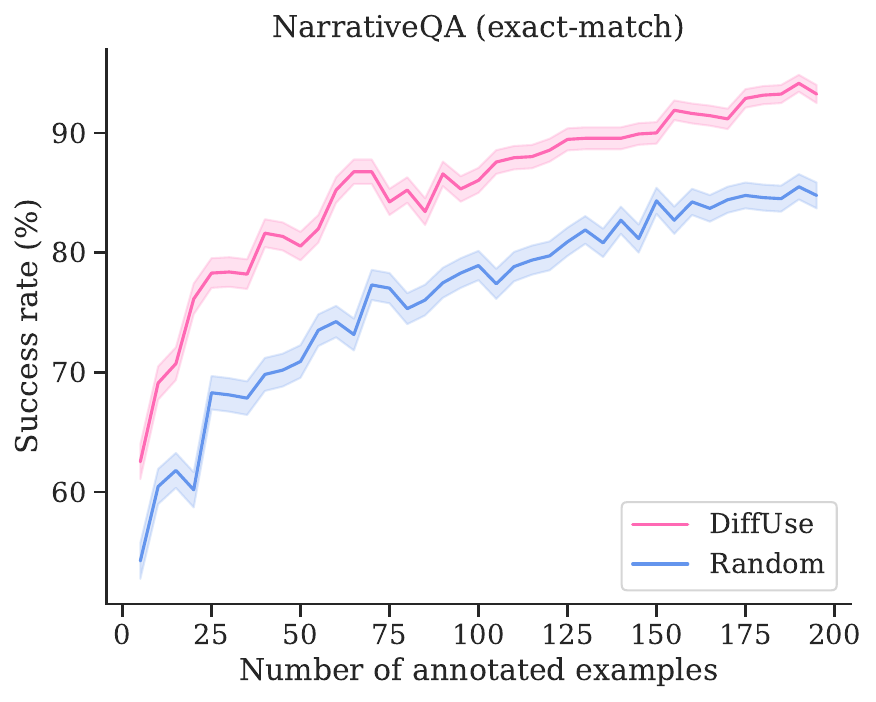}}
    \subfloat{\includegraphics[width=0.32\textwidth]{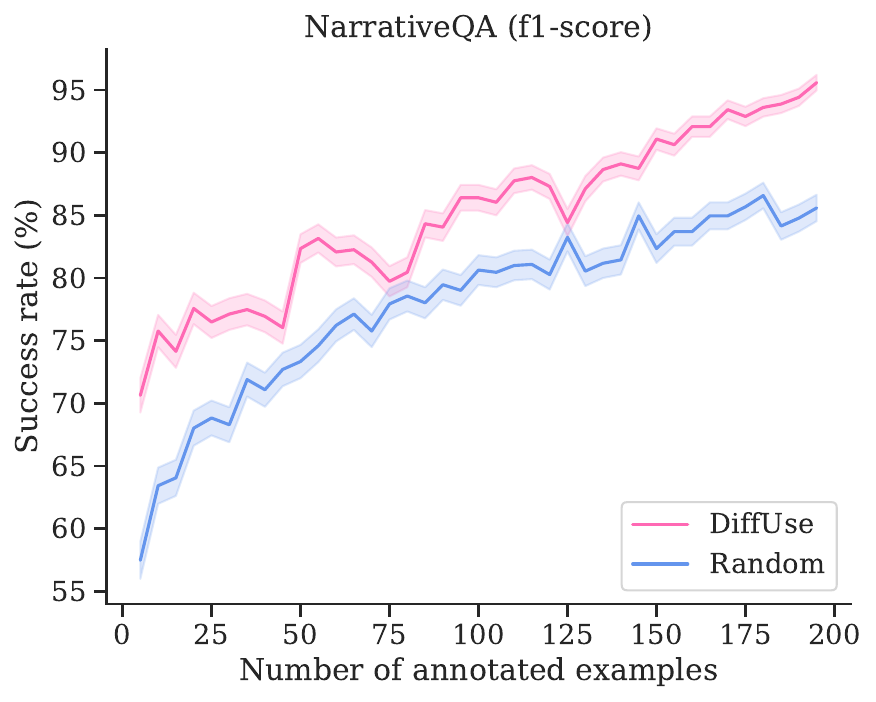}}
    \subfloat{\includegraphics[width=0.32\textwidth]{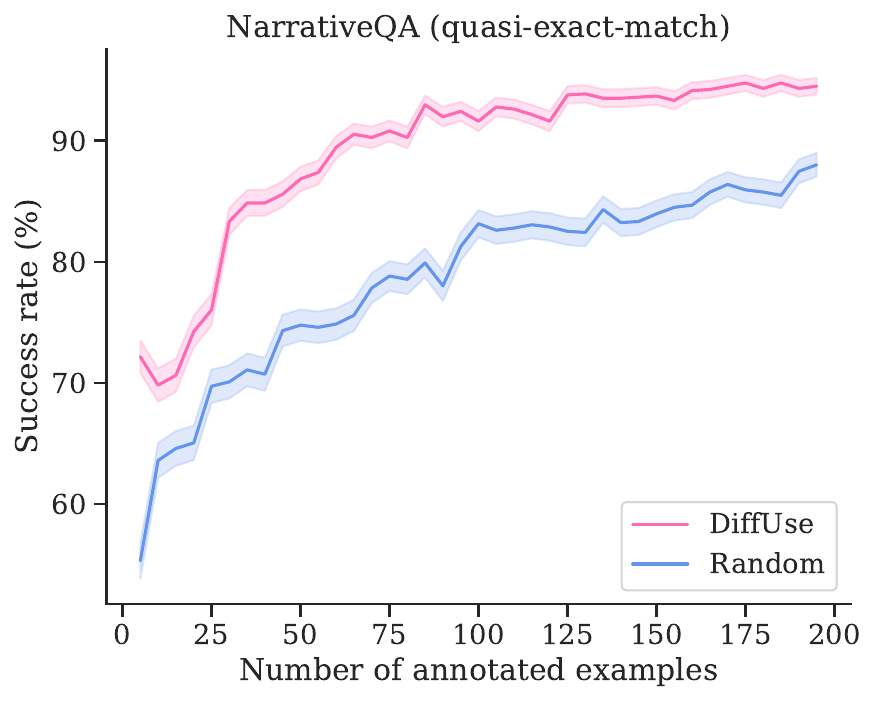}}\\[-2ex]
    \subfloat{\includegraphics[width=0.32\textwidth]{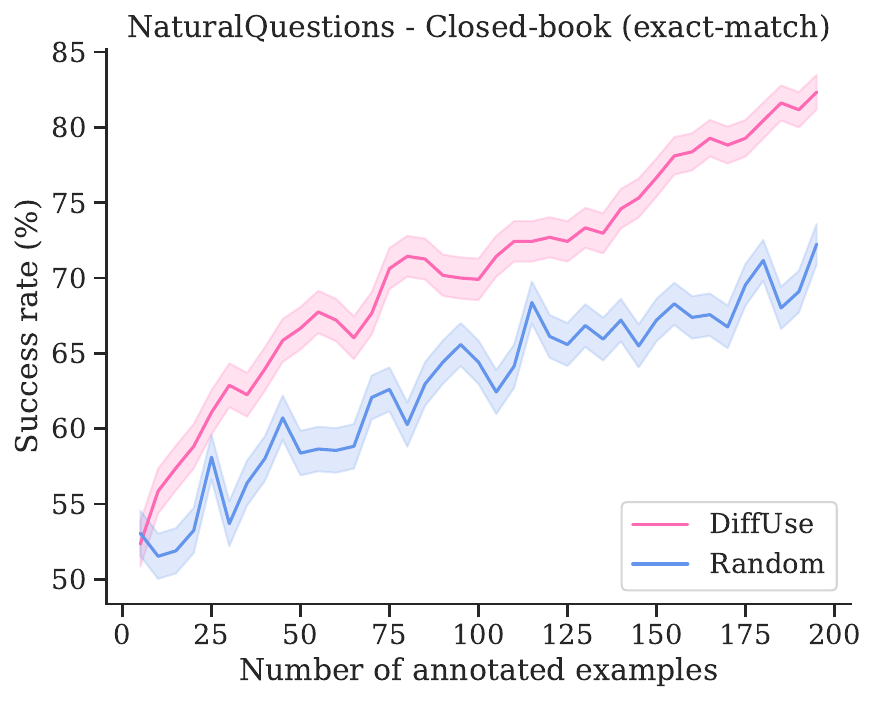}}
    \subfloat{\includegraphics[width=0.32\textwidth]{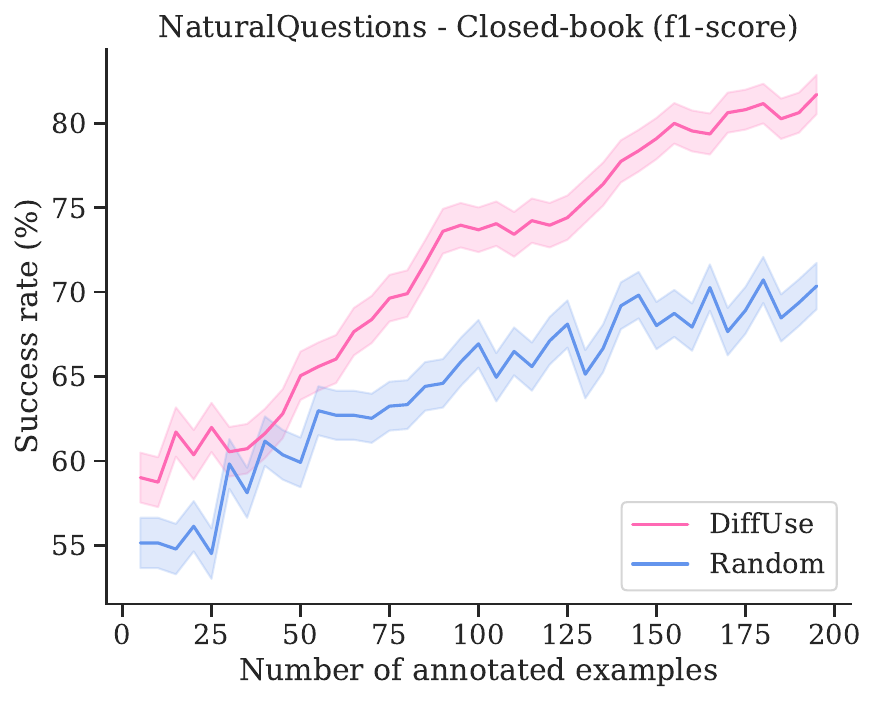}}
    \subfloat{\includegraphics[width=0.32\textwidth]{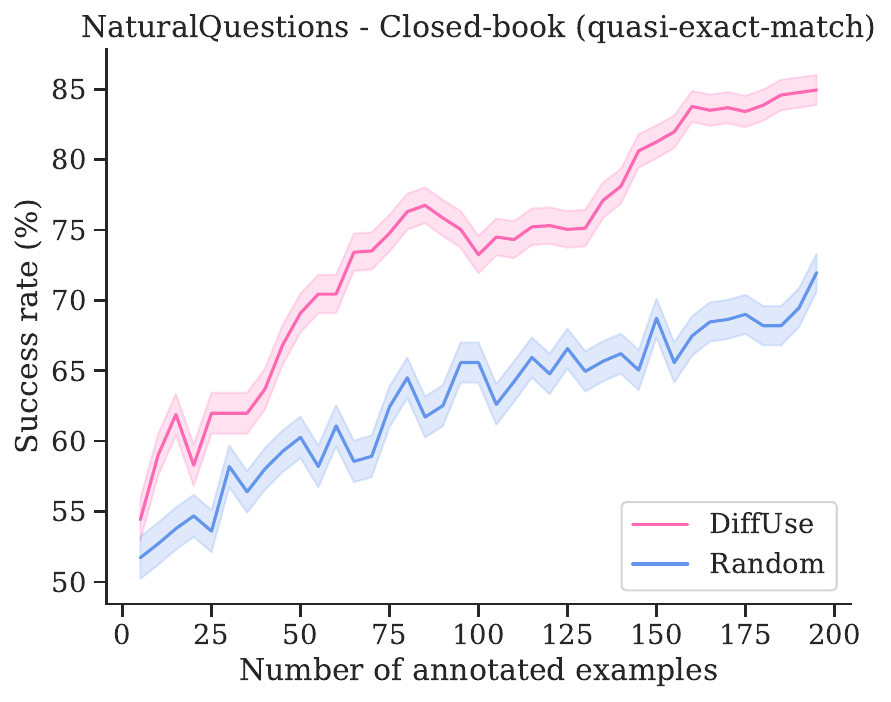}}\\[-2ex]
    \subfloat{\includegraphics[width=0.32\textwidth]{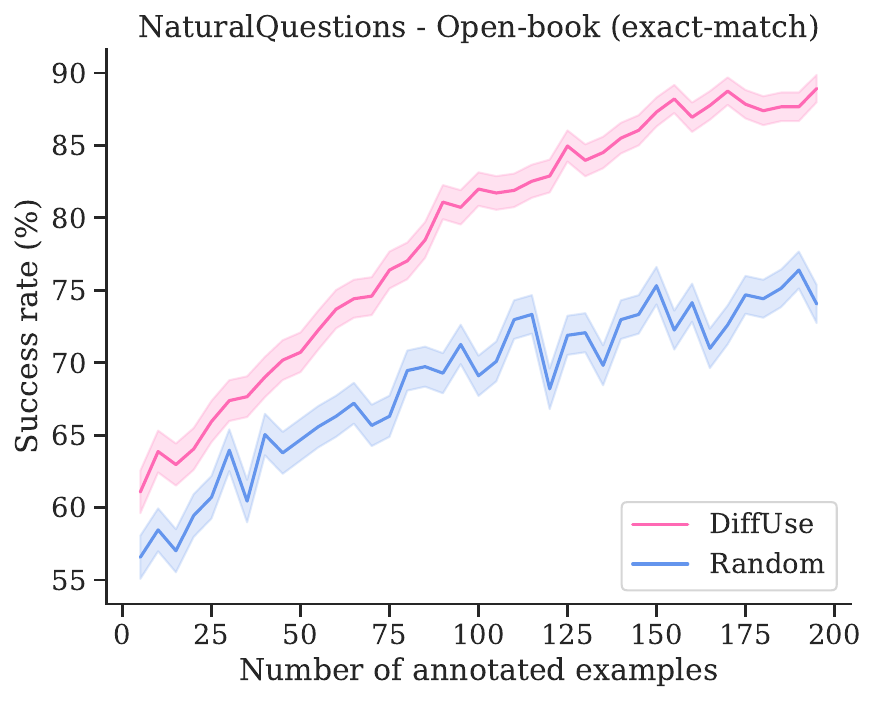}}
    \subfloat{\includegraphics[width=0.32\textwidth]{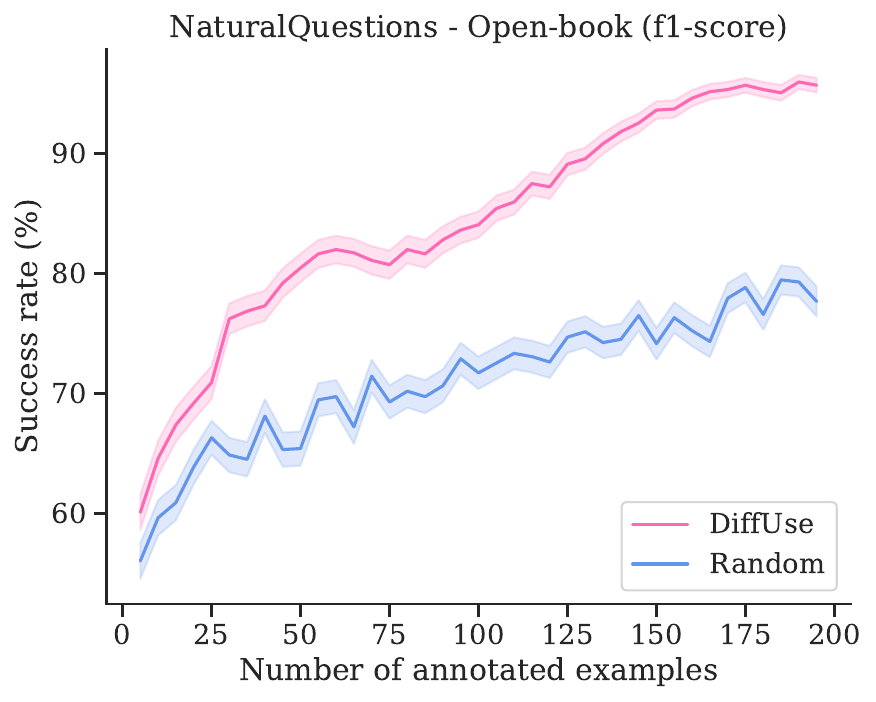}}
    \subfloat{\includegraphics[width=0.32\textwidth]{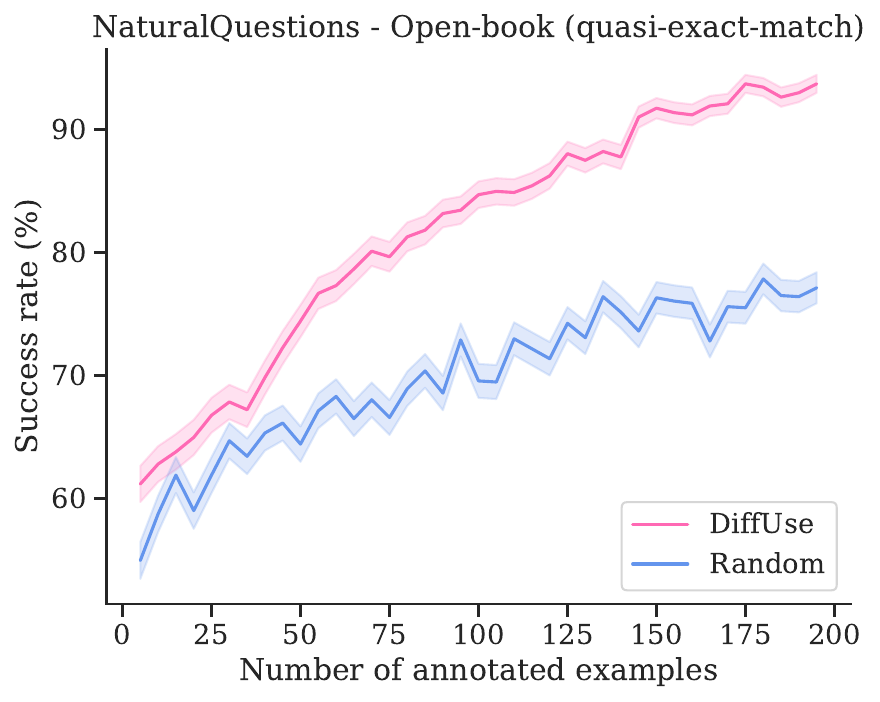}}\\[-2ex]
    \subfloat{\includegraphics[width=0.32\textwidth]{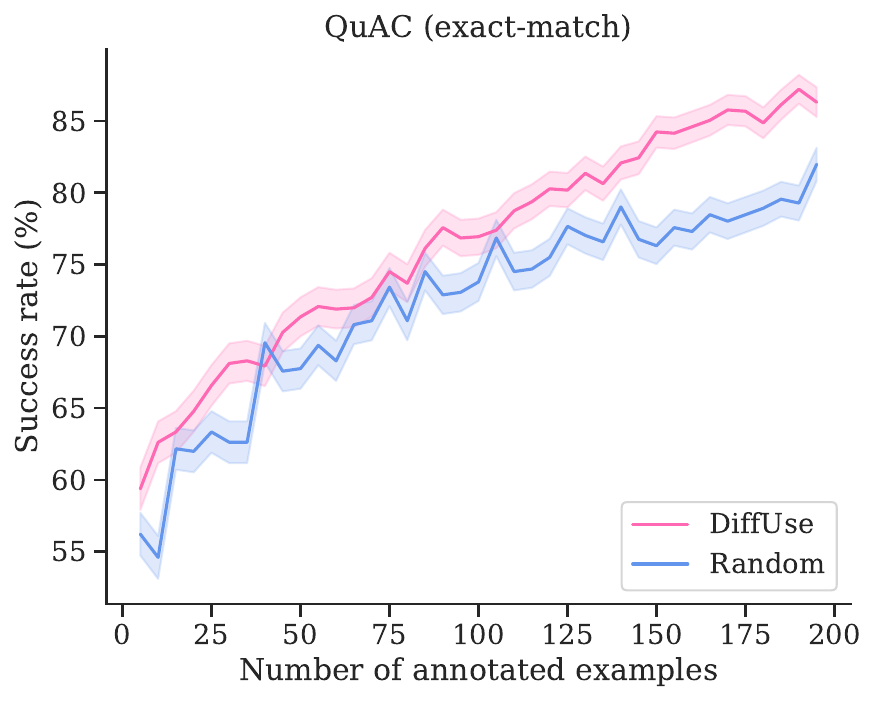}}
    \subfloat{\includegraphics[width=0.32\textwidth]{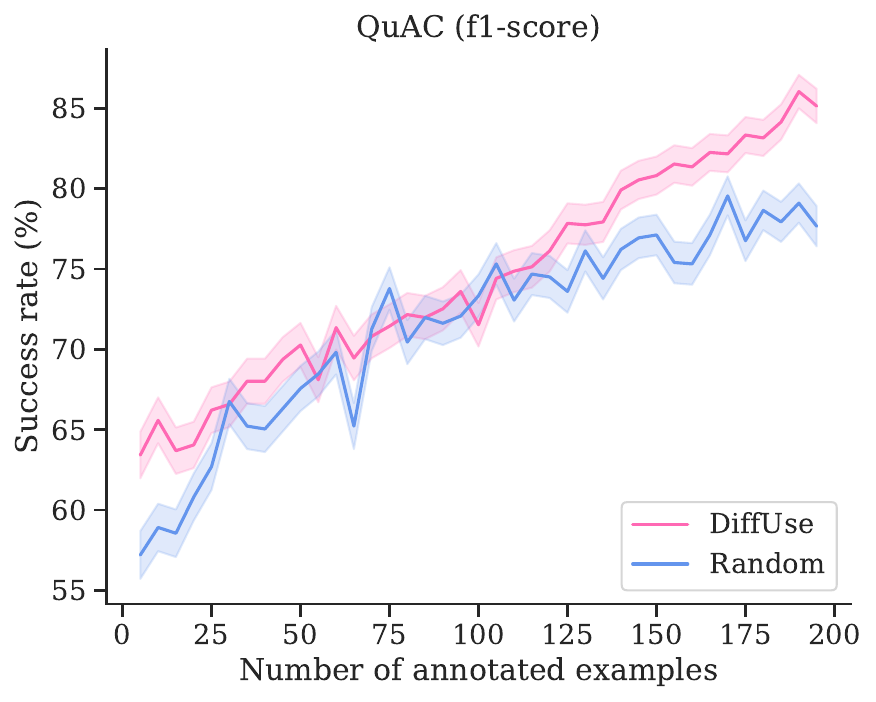}}
    \subfloat{\includegraphics[width=0.32\textwidth]{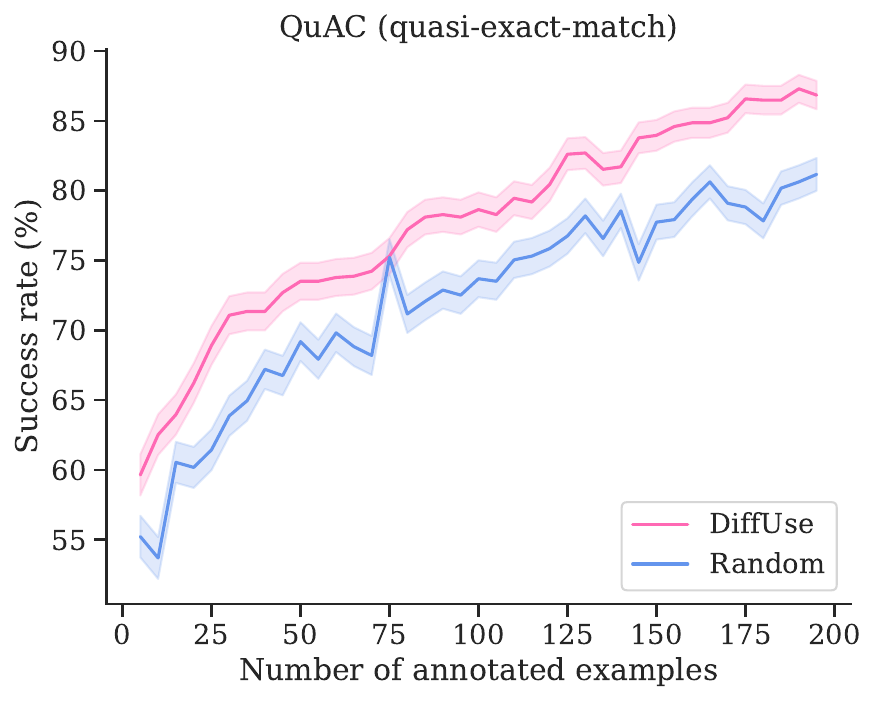}}\\[-2ex]
    \subfloat{\includegraphics[width=0.32\textwidth]{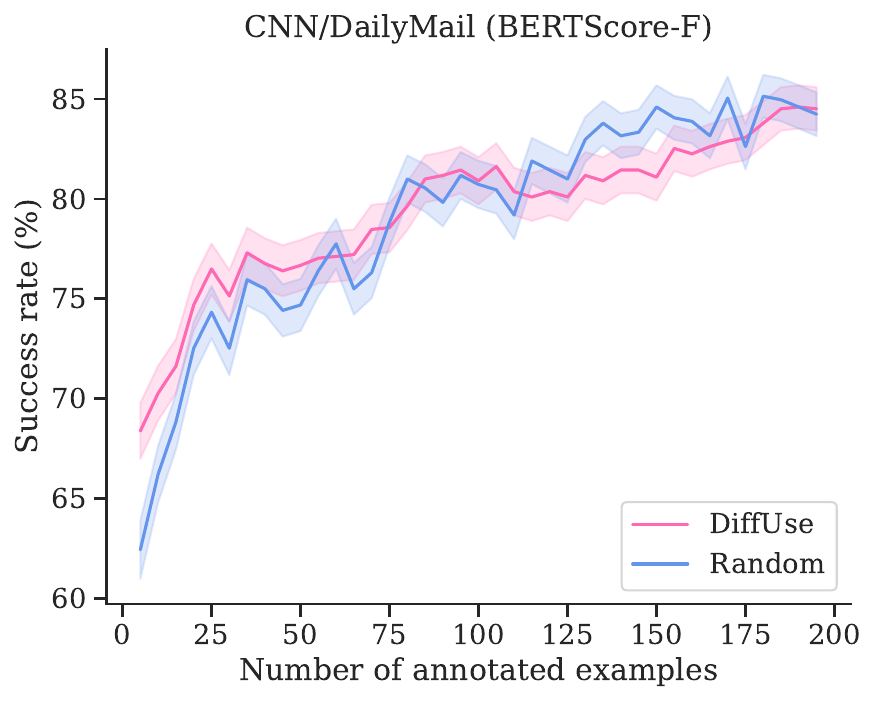}}
    \subfloat{\includegraphics[width=0.32\textwidth]{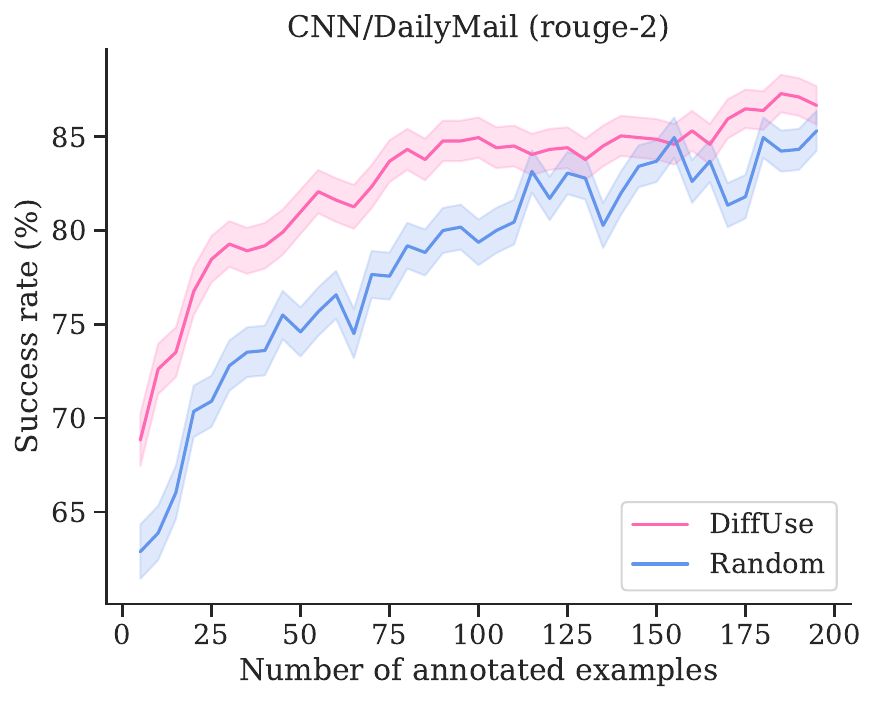}}
    \subfloat{\includegraphics[width=0.32\textwidth]{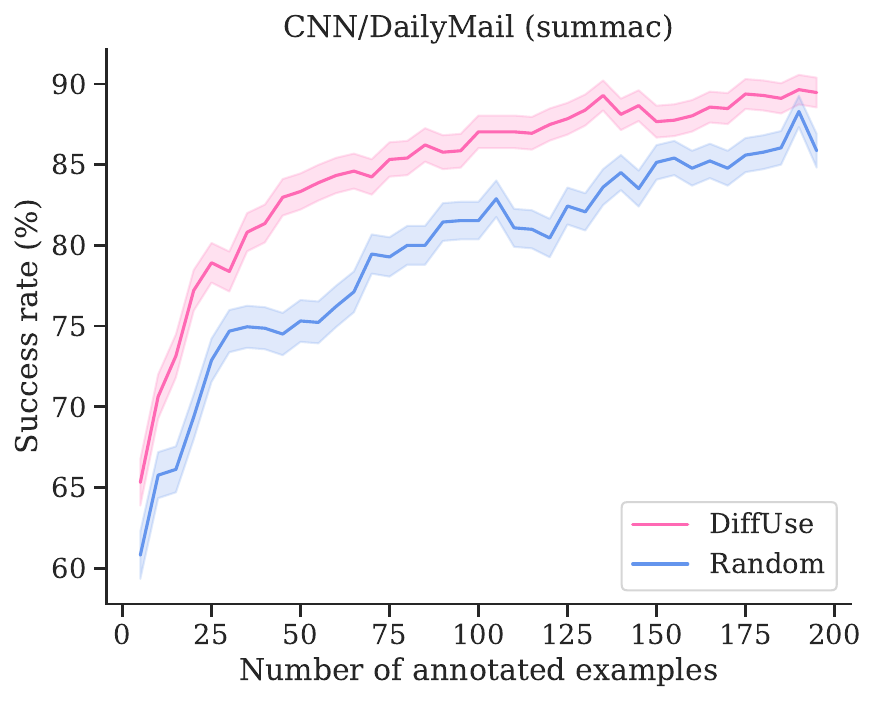}}\\[-2ex]
    \subfloat{\includegraphics[width=0.32\textwidth]{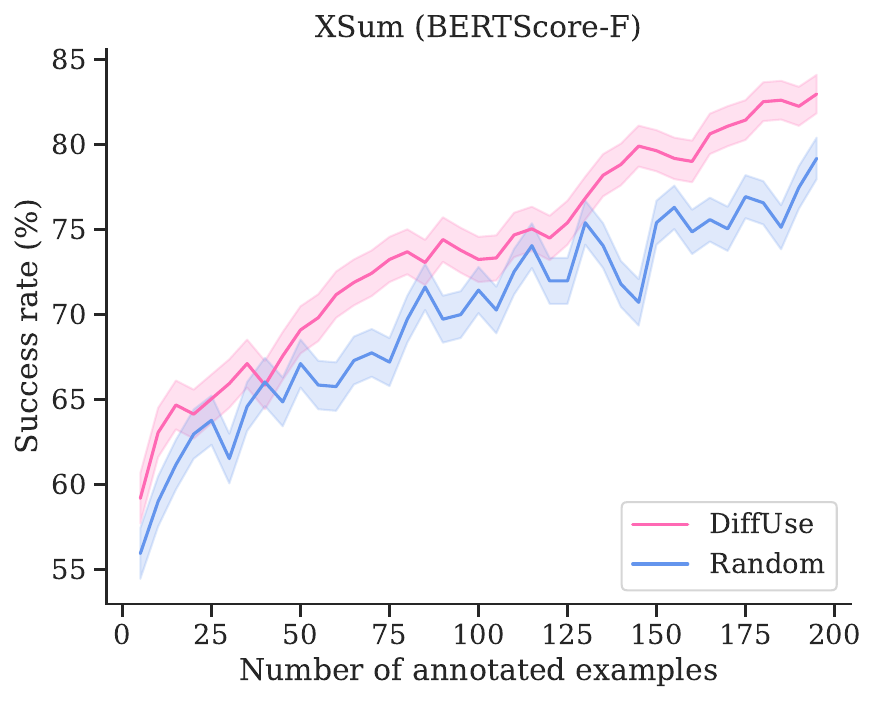}}
    \subfloat{\includegraphics[width=0.32\textwidth]{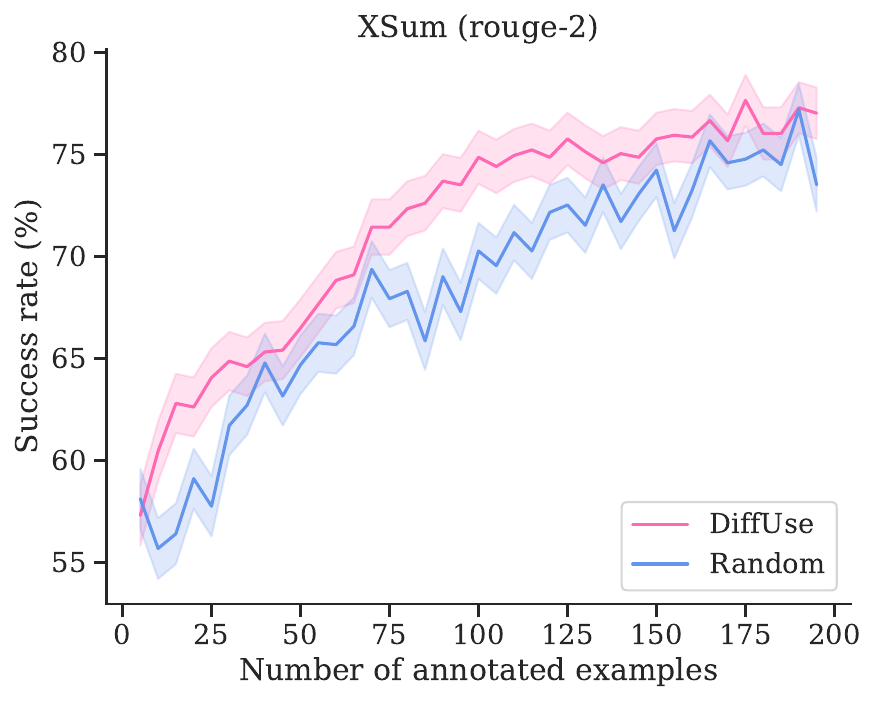}}
    \subfloat{\includegraphics[width=0.32\textwidth]{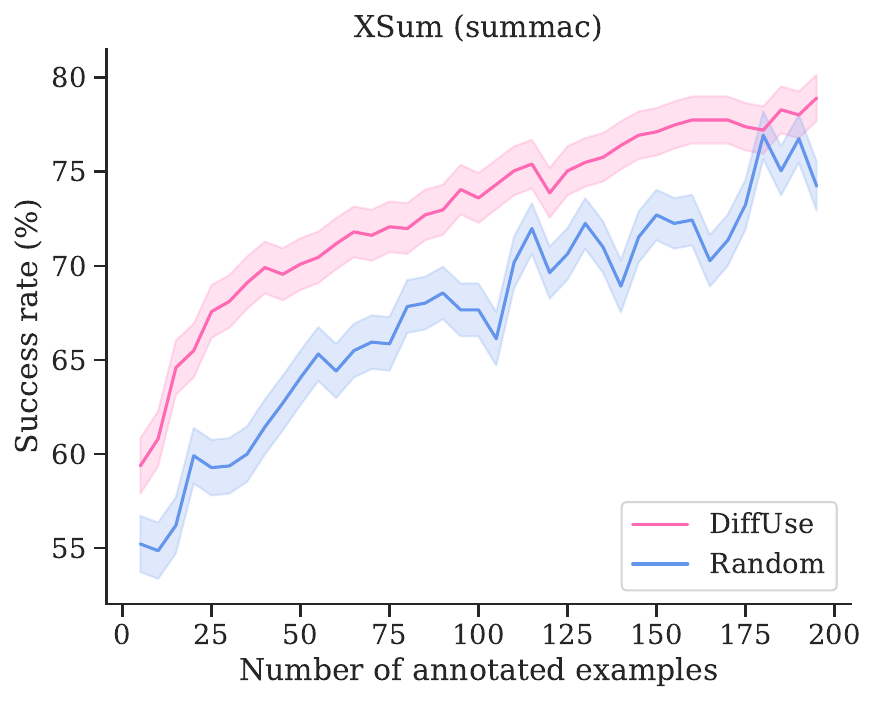}}\\[-2ex]
\caption{\textbf{Prompts results}. Plots depict success rates of prompt preference estimation, aggregated over $111$ unique pairs. Each panel depicts a different combination of dataset and "oracle" (reference-based evaluation metric).}
\label{fig_prompts}
\end{figure*}